%% file: main.tex
\documentclass{article}
\usepackage{template/PRIMEarxiv}

\usepackage[utf8]{inputenc}
\usepackage[T1]{fontenc}
\usepackage{hyperref}
\usepackage{url}
\usepackage{booktabs}
\usepackage{amsfonts}
\usepackage{nicefrac}
\usepackage{microtype}
\usepackage{lipsum}
\usepackage{fancyhdr}
\usepackage{graphicx}
\graphicspath{{media/}}

\title{The RL/LLM Taxonomy Tree: Reviewing Synergies Between Reinforcement Learning and Large Language Models
}

\author{
  Moschoula Pternea\\
  Microsoft\\
  \texttt{mpternea@microsoft.com}
  \\\And
  Prerna Singh\\
  Microsoft\\
  \texttt{prernasingh@microsoft.com}
  \\\And
  Abir Chakraborty\\
  Microsoft\\
  \texttt{abir.chakraborty@microsoft.com}
  \\\And
  Yagna Oruganti\\
  Microsoft\\
  \texttt{yaorugan@microsoft.com}
  \\\And
  Mirco Milletari\\
  Microsoft\\
  \texttt{mimillet@microsoft.com}
  \\\And
  Sayli Bapat\\
  Microsoft\\
  \texttt{saylibapat@microsoft.com}
  \\\And
    Kebei Jiang\\
  Microsoft\\
  \texttt{kebei.jiang@microsoft.com}
}

\usepackage{booktabs}
\usepackage{multirow}
\usepackage[numbers]{natbib} 
\usepackage{tabularx}
\usepackage{lscape} 
\usepackage{amsmath}
\usepackage{tikz}
\usepackage{tabularray}

\def\checkmark{\tikz\fill[scale=0.4](0,.35) -- (.25,0) -- (1,.7) -- (.25,.15) -- cycle;} 

\makeatletter
\newcommand\mytuple[1]{%
  \@tempcnta=0
  \bigl\langle
  \@for\@ii:=#1\do{%
    \@insertbreakingcomma
    \textit{\@ii}
  }%
  \bigr\rangle
}
\def\@insertbreakingcomma{%
  \ifnum \@tempcnta = 0 \else\,,\ \linebreak[1] \fi
  \advance\@tempcnta\@ne
}
\makeatother
\usepackage{xurl} 

\begin{document}

\maketitle
\author{\name Moschoula Pternea \email mpternea@microsoft.com \\
       \name Prerna Singh \email prernasingh@microsoft.com \\
       \name Abir Chakraborty \email abir.chakraborty@microsoft.com \\
       \name Yagna Oruganti \email yaorugan@microsoft.com \\
       \name Mirco Milletari \email mimillet@microsoft.com \\
       \name Sayli Bapat \email saylibapat@microsoft.com \\
       \name Kebei Jiang \email kebei.jiang@microsoft.com \\
       \addr One Microsoft Way, Redmond, WA 98052
   }

\maketitle
\begin{abstract}
In this work, we review research studies that combine Reinforcement Learning (RL) and Large Language Models (LLMs), two areas that owe their momentum to the development of deep neural networks. We propose a novel taxonomy of three main classes based on the way that the two model types interact with each other. The first class, \texttt{RL4LLM}, includes studies where RL is leveraged to improve the performance of LLMs on tasks related to Natural Language Processing. \texttt{RL4LLM} is divided into two sub-categories depending on whether RL is used to directly fine-tune an existing LLM or to improve the prompt of the LLM. In the second class, \texttt{LLM4RL}, an LLM assists the training of an RL model that performs a task that is not inherently related to natural language. We further break down LLM4RL based on the component of the RL training framework that the LLM assists or replaces, namely reward shaping, goal generation, and policy function. Finally, in the third class, \texttt{RL+LLM}, an LLM and an RL agent are embedded in a common planning framework without either of them contributing to training or fine-tuning of the other. We further branch this class to distinguish between studies with and without natural language feedback. We use this taxonomy to explore the motivations behind the synergy of LLMs and RL and explain the reasons for its success, while pinpointing potential shortcomings and areas where further research is needed, as well as alternative methodologies that serve the same goal.
\end{abstract}

\input{SECTIONS/1_INTRODUCTION}
\input{SECTIONS/2_BACKGROUND-SCOPE-CONTRIBUTIONS}
\input{SECTIONS/3_TAXONOMY}
\input{SECTIONS/4_RL4LLM}

\input{SECTIONS/5_LLM4RL}
\input{SECTIONS/6_RLPLUSLLM}

\input{SECTIONS/7_DISCUSSION}
\input{SECTIONS/8_CONCLUSIONS}

\bibliographystyle{abbrvnat}
\bibliography{references}

\end{document}

%% file: SECTIONS/1_INTRODUCTION.tex
\section{Introduction} \label{sec:intro}

Reinforcement Learning (RL) and Large Language Models (LLMs) are experiencing tremendous progress over recent  years, with the common factor behind the growth of both Artificial Intelligence domains being the development of Deep Neural Networks (DNNs).

The foundations of Markov Decision Processes (MDPs), which are at the core of every RL model, can practically be traced back to the mid-20th century \citep{bellman1957}, when they originated in the field of stochastic control \citep{howard1960} with the goal to model sequential decision making in uncertain environments. Reinforcement Learning proposed a formal framework for approaching sequential decision making problems by adapting concepts from behavioral psychology, where an agent can learn by interacting with their environment and utilizing their past experience \citep{suttonbarto,franccois2018introduction}. However, it was the development of Deep Reinforcement Learning \citep{franccois2018introduction} that addressed the key challenges of traditional value and policy function approximations by tackling the curse of dimensionality through efficient state representation, better generalization, and sample efficiency. As a result, Deep RL algorithms have become increasingly popular over recent years, with applications in control systems, robotics, autonomous vehicles, healthcare, finance, to name only a few.

Similarly, Natural Language Processing (NLP) problems, like speech recognition, natural language understanding, machine translation, text summarization, etc., have long been successfully solved by machine learning algorithms, ranging from Naïve Bayes and Maximum Entropy Models to Decision Trees and Random Forests. \cite{nlpreviewjumping} attributed the impressive development of NLP methods to three overlapping curves – Syntactics, Semantics, and Pragmatics – and foresaw the eventual evolution of NLP research to natural language understanding. Deep learning revolutionized NLP tasks with various Neural Network architectures, such as Recurrent Neural Networks (RNNs), Long Short-Term Memory (LSTM), Convolutional Neural Networks (CNN) and, more recently, Transformers \citep{vaswani2017attention}. Eventually, Deep Neural Networks are opening new avenues in the field of Natural Language Processing with the development of LLMs, which are language models trained on massive amounts of text using specialized hardware, like GPU and TPUs, to perform complicated NLP tasks.

Apart from owing their growth to the development of Deep Neural networks, LLMs and RL are intertwined from a theoretical and practical perspective because  they can both be formulated and approached as sequential modeling problems: LLMs generate text in a sequential decision-making framework, selecting the most likely next word or phrase. As noted by \cite{ramamurthy2023reinforcement}, ``if we view text generation as a sequential decision-making problem, reinforcement learning (RL) appears to be a natural conceptual framework’’. On the other side, RL deals inherently with control problems, where an agent must select the most appropriate action, interact with its environment, observe the result of its action, and continue in a loop of state-observation-action-reward for a possibly infinite horizon.

Motivated by the impressive prominence of Reinforcement Learning and Large Language Models, along with the impressive range of practical applications they both present, we perform a comprehensive literature review of studies that embed Large Language Models and Reinforcement Learning Agents in a common computational framework. More specifically, we are proposing a novel taxonomy to classify those studies based on the way that the LLM and the RL agent interact in the framework. With this taxonomy as the backbone of our review, we break down the state-of-art frameworks into their fundamental components and present details to describe the ways in which the two model types collaborate in each study. In parallel, we explain the key motivational factors and the reasons behind the success of this collaboration. We also review the potential limitations of this synergy and present alternative state-of-art methods that, while not parts of this taxonomy, have been developed with the intent to address the same issues as the studies that we are focusing on. This thorough categorization will help researchers obtain a better understanding of the dynamics of this synergy, explain trends and opportunities in the intersection of RL and LLMs, and serve as a starting point for the development of novel AI frameworks that combine the best of both these worlds.

The rest of this paper is structured as follows: in section \ref{sec:background-contributions}, we provide the fundamental terms and concepts around Reinforcement Learning, transformers, and LLMs, to facilitate the reader in understanding the material that follows, and outline the scope and contributions of this study. In section \ref{sec:taxonomy}, we provide an overview of the proposed taxonomy. Sections \ref{sec:rl4llm}, \ref{sec:llm4rl} and \ref{sec:rlplusllm} are dedicated to the main classes of our proposed taxonomy, corresponding to \texttt{RL4LLM}, \texttt{LLM4RL}, and \texttt{RL+LLM}, respectively. In section \ref{sec:discussion}, we discuss the emerging patterns of this taxonomy and the reasons behind the success of this synergy, as well as shortcomings and alternative methods to achieve the same goals. Finally, in section \ref{sec:conclusions} we summarize our findings and conclusions and propose paths for future research.

%% file: SECTIONS/2_BACKGROUND-SCOPE-CONTRIBUTIONS.tex
\section{Background, State-of-Art, Scope, and Contributions} \label{sec:background-contributions}

\subsection{Overview of RL and LLMs} \label{subsec:overview}

Reinforcement learning encompasses a range of algorithms created to tackle problems that require a series of decisions and it differs significantly from both supervised and unsupervised learning methods: it requires the learning system, also referred to as an agent, to independently determine the best sequence of actions to achieve its goal through interaction with its environment. Reinforcement methods are primarily divided into three categories: dynamic programming, Monte Carlo methods, and temporal difference methods. All these methods present the decision-making issue as a Markov decision process (MDP), a mathematical approach to solving sequential decision-making problems that involves a state set $S$, an action set $A$, a transition function $T$, and a reward function $R$. The goal of an MDP $(S, A, T, R)$ is to determine an optimal policy function $\pi$, which outlines the agent's behavior at any given time. Essentially, a policy maps the set of states $S$ perceived from the environment to a set of actions $A$ that should be performed in those states. The objective of the agent is to maximize a cumulative reward $r\in R$ by selecting the actions to perform in each state $s$. When in state $s$, the agent performs action $a$, receives reward $r$ from its environment, and then moves to the next state $s'$. Each step can therefore be represented as a transition tuple $(s, a, r, s')$. The process to estimate the policy $\pi$ depends on the algorithm in use and the specifics of the problem. In certain instances, especially when the state and action space are tractable, the policy can be stored as a lookup table, while in others, a function approximator (such a neural network) is used.

Within the realm of Natural Language Processing (NLP), Large Language Models (LLMs) have become a ubiquitous component. The primary role of a language model is to establish a probability distribution across word sequences, a process achieved by the application of the chain rule of probability to dissect the joint probability of a word sequence into conditional probabilities. Language models may be unidirectional, forecasting future words based on past ones as seen in n-gram models, or bidirectional, making predictions for a word based on both antecedent and subsequent words as exemplified by Transformer models. Owing to advancements in deep learning, neural language models have seen a surge in popularity. An LLM makes use of a specific kind of neural network known as a transformer \citep{vaswani2017attention}, which uses a mechanism called attention to weigh the influence of different words when producing an output. The term ``large'' in this context signifies the substantial number of parameters these models hold. LLMs are capable of responding to queries, authoring essays, summarizing texts, translating languages, and even generating poetry. Some of the most popular LLMs include BERT \citep{devlin2019bert}, GPT \citep{brown2020language}, PaLM \citep{chowdhery2022palm}, and LaMDA \citep{thoppilan2022lamda}.

\subsection{State-of-Art Review Studies} \label{subsec:back}

As rapidly advancing fields, with a wide range of applications, both Reinforcement Learning and Natural Language Processing have been the focus of numerous studies that aim to synthesize and evaluate state-of-art research in each area.

Since it first emerged, RL has been of particular interest to researchers in computer science, robotics, and control and, as a result, numerous surveys on RL have been published, ranging from general overview of RL \citep{kaelbling1996reinforcement,arulkumaran2017brief} to comprehensive reviews that focus on a particular technique (Offline RL, \citep{prudencio2023survey}; Meta-Reinforcement Learning, \citep{beck2023survey}; RL on graphs, \citep{nie2023reinforcement}; Evolutionary RL, \citep{bai2023evolutionary}; Hierarchical RL, \citep{shubhan2021hierarchical}; Multi-Agent Deep RL, \citep{gronauer2022multi,du2021survey}, application (healthcare, \citep{yu2021healthcare}; robotics, \citep{garaffa2023robotics}; combinatorial optimization, \citep{mazyavkina2021reinforcement};  generative AI, \citep{cao2023reinforcement}, learning assumptions (dynamically varying environment, \citep{padakandla2021dynamically}. Owing to the rapid emergence of LLMs, we are also beginning to witness review papers dedicated to this topic, like the comprehensive review on RLHF by \cite{sun2023reinforcement}.

A similar trend can be observed in Natural Language Processing, with numerous examples of survey studies providing an overall study of the concepts and methods in the field \citep{torfi2020natural}, particularly since the introduction of deep learning for NLP \citep{otter2020survey,torfi2020natural,chowdhary2020natural}. Similarly to the case of RL, literature review studies specialize by application (e.g., healthcare, \citep{wu2022survey}; fake news detection, \citep{oshikawa2018survey}; bioinformatics, \citep{zeng2015survey} or focus on particular methods (pretrained models, \citep{qiu2020pre}, graphs, \citep{nastase2015survey}, etc.

Not surprisingly, LLMs themselves, which lie at the intersection of Natural Language Processing and Reinforcement Learning, have attracted the attention of researchers worldwide, resulting already in an impressive wealth of comprehensive literature review publications, ranging from general reviews \citep{zhao2023survey,xi2023rise,min2021recent,yang2023harnessing,liu2023gpt} to surveys focusing on different aspects of LLMs, like evaluation \citep{guo2023evaluating,chang2023survey}, alignment with humans \citep{shen2023large,wang2023aligning}, explainability \citep{zhao2023explainability}, Responsible AI considerations \citep{gallegos2023bias}, a knowledge acquisition and updating \citep{cao2023life,wang2023knowledge,pan2023unifying} as well as using LLMs for specific applications like information retrieval \citep{zhu2023large}, natural language understanding \citep{du2023shortcut}, instruction tuning \citep{zhang2023instruction}, software engineering \citep{wang2023software,fan2023large}, recommendation systems \citep{wu2023survey,li2023large,lin2023recommender}, opinion prediction \citep{kim2023aiaugmented}, and other applications.

As will be explained in more detail in subsection \ref{subsec:scope}, this survey examines Reinforcement Learning and Large Language Models from a completely different angle compared to the aforementioned review papers, since it focuses exclusively on studies where RL and LLMs are both indispensable components of the same computational framework.

\subsection{Scope of This Study} \label{subsec:scope} 
As explained in section \ref{sec:intro}, we are presenting a survey on studies that combine Reinforcement Learning and Large Language Models in a common modeling framework and we are proposing a mew taxonomy to classify them. The taxonomy is visualized as the RL/LLM Taxonomy Tree (Fig. \ref{fig_tree}), which maps each study to a tree node, according to the details of the synergy between the two models. 

 Although Reinforcement Learning – in its RLHF form – is an essential component of any Large Language Model, our review is only concerned with studies that involve already trained LLMs, which are then either improved and fine-tuned with RL - beyond RLHF, that was used to train the original model - or combined with some RL agent to perform a downstream task. Studies where Reinforcement Learning is limited to training the original LLM are beyond the scope of our taxonomy.

In addition, literature exhibits state-of-art survey papers that focus on the use of LLMs for tasks that are not related to natural language, including the augmentation of LLMs with reasoning and other skills \citep{huang2022towards,mialon2023augmented,wei2022emergent}, multimodal LLMs \citep{wang2023largescale}, and autonomous agents \citep{wang2023survey,liu2023agentbench}. While in this survey we are also reviewing studies where LLMs are used to perform general tasks (sections \ref{sec:rl4llm} and \ref{sec:rlplusllm}, we are exclusively focusing on those where the RL agent, rather that an LLM, is performing a downstream task, and the LLM assists the framework either at training (\texttt{LLM4RL} class, section \ref{sec:llm4rl} or at inference (\texttt{RL+LLM} class, section \ref{sec:rlplusllm}). Therefore, contrary to \cite{liu2023agentbench} and \cite{wang2023survey}, we are not concerned with evaluating the performance of the LLMs as autonomous agents. For the sake of completeness, we still discuss the use of LLMs to tasks that are not related to Natural Language, along with Multimodel LLMs in section \ref{sec:discussion}.

Finally, the use of pretrained language models to aid RL agents through reward design \citep{cao2023temporal}, policy priors \citep{choi2022lmpriors} or policy transfer \citep{jiang2023languageinformed} preceded the development of LLMs. While we refer to those studies for the sake of completeness, our taxonomy and subsequent analysis only captures those studies where the language model used is an LLM.

\subsection{Contributions of This Study}

To the best of our knowledge, this is the first study that attempts a thorough review of the state-of-art research on the intersection of Large Language Models and Reinforcement Learning. To this direction, we have identified 24 publications that combine LLMs and RL and fall within the scope of this review as described in subsection \ref{subsec:scope}. Our goal is to examine how the two distinct models are embedded in a common framework to achieve a specific task. A set of common patterns emerged from the review of those studies, which helped us categorize the studies according to the way that the two models collaborate, as well as the nature of the end task to be achieved. Therefore, we are proposing a novel, systematic taxonomy of those studies that helps researchers understand the scope and applications of the various synergies between RL and LLMs. First, we follow the proposed taxonomy to individually present the key features, goals, and highlights of each study that we have reviewed. Then, we shift our focus to obtaining a global perspective on the collective goals of each study category and explain their strengths and potential shortcomings.

In summary, the contribution of our work is threefold:
\begin{enumerate}
\item We collect, review, and analyze state-of-art studies which combine Reinforcement Learning  and Large Language Models in the same framework. 
\item We propose a novel taxonomy to explain the synergy between RL and LLMs. In particular, we visualize the classification of the RL/LLM studies using the RL/LLM Taxonomy Tree, which includes three main classes which collectively capture any problem that utilizes both RL and LLMs. The criterion for generating the three classes is whether RL is utilized to improve the performance of an LLM (class 1 – \texttt{RL4LLM}), or an LLM is used to train an RL agent to perform a non-NLP task (class 2 – \texttt{LLM4RL}), or whether the two models are trained independently and then embedded in a common framework to achieve a planning task (class 3 – \texttt{RL+LLM}). The order in which we present the individual studies (contribution 1) is based on this taxonomy.
\item We utilize our findings from the taxonomy to discuss the applications of this synergy, explain the reasons for its success, identify strengths and potential weaknesses, and investigative alternative ways towards achieving the same tasks.
\end{enumerate}

%% file: SECTIONS/3_TAXONOMY.tex
\section{The RL/LLM Taxonomy Tree} \label{sec:taxonomy}

Even though LLMs are an emerging field, there exists a substantial body of literature dedicated to their intersection with Reinforcement Learning. We can readily discern a top-level classification based on the interplay between the two models - RL agent and LLM – as the key classification criterion. We have therefore identified the following classes of studies, which constitute the core classes of our taxonomy:

\begin{figure*} \label{fig_tree}
\centering
\centerline{\includegraphics[width=\textwidth]{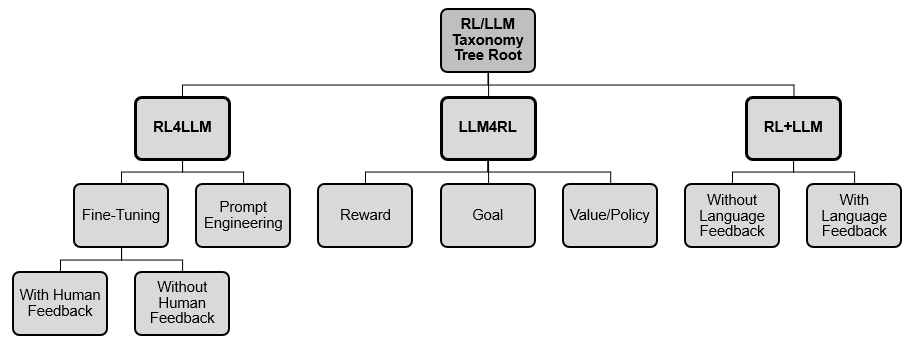}}
\caption{The RL/LLM Taxonomy Tree.}
\end{figure*}

\begin{enumerate}
\item \texttt{RL4LLM}. These studies use RL to improve the performance of the LLM in an NLP task.
\item \texttt{LLM4RL}. These studies use an LLM to supplement the training of an RL model that performs a general task that is not inherently related to natural language.
\item \texttt{RL+LLM}. These studies combine RL models with LLM models to plan over a set of skills, without using either model to train or fine-tune the other.
\end{enumerate}

The frameworks belonging to \texttt{RL4LLM} class start from a trained LLM and subsequently utilize RL to modify it, with the goal to improve its performance on specific tasks or align it to user intent and ethical AI standards. On the contrary, studies in \texttt{LLM4RL} category utilize the LLM as a component of an RL training framework with the goal of helping an RL agent perform a specific task. Finally, \texttt{RL+LLM} involves the two models as independent components of a common framework, without either of them directly participating in the training or tuning of the other in any way.

Interestingly, the way that the RL agent and the LLM types interact in each case is directly tied to the goal of the synergy, which helps us identify a mapping between the structure of each framework and its end goal. More specifically, \texttt{RL4LLM} studies aim to improve the performance of the LLM in a downstream task that is related to natural language processing, such as text summarization, question-answering, or conversation. In turn, the goal of \texttt{LLM4RL} frameworks is to improve the training efficiency or performance of a control task that would still rely on RL in the absence of the LLM and is therefore generally not related to NLP. Finally, studies of \texttt{RL+LLM} generally use the LLM to plan over individual skills that have been learned through RL.

Studies within each top-level class of the RL/LLM Taxonomy Tree exhibit significant variety in the way that the RL agent and the LLM interact in each framework. thus requiring further refinement of the taxonomy. Specifically, studies within \texttt{RL4LLM} can be broken down into the following subcategories:
\begin{itemize}
\item \texttt{RL4LLM-Fine-tuning}: Encompasses studies where RL is used to perform model fine-tuning, which involves tweaking the model parameters, until the model achieves the desired performance. This subclass can be further refined according to the presence or absence of human feedback.
\item \texttt{RL4LLM-Prompt Engineering}: Includes studies where RL is used to iteratively update the prompt of the LLM, until the model achieves the desired performance.
Similarly, \texttt{LLM4RL} can be further divided according to the component of the RL framework that is assisted, replaced, or represented by the LLM, namely:
\item \texttt{LLM4RL-Reward}: Includes studies where the LLM is used to design the reward function of the RL agent.
\item \texttt{LLM4RL-Goal}: includes studies where the LLM is utilized for goal setting, which applies to goal-conditioned RL settings.
\item \texttt{LLM4RL-Policy}: includes studies where the LLM represents the policy function to be learned, or directly assists its training or pretraining.
\end{itemize}
Finally, \texttt{RL+LLM} class is branched into two subclasses:
\begin{itemize}
\item \texttt{RL+LLM-No Language Feedback}: studies where the prompt of the LLM is updated during the planning process.
\item \texttt{RL+LLM-With Language Feedback}: studies where the prompt of the LLM stays fixed throughout the planning process.
\end{itemize}

The RL/LLM Taxonomy Tree is visualized in Fig. \ref{fig_tree}, while Table \ref{tab:studies} maps the research studies to the particular subclasses they belong to, corresponding to the leaf nodes of the RL/LLM Taxonomy Tree.

To the best of our knowledge, the classification proposed by the RL/LLM Taxonomy Tree is exhaustive and captures all state-of-art studies that fall within the scope of our taxonomy (\ref{subsec:scope}) as of today. We have so far identified 24 studies that fall under the scope of this review and can direcly be mapped to RL/LLM Taxonomy Tree leaves. Therefore, it has the potential to serve as a reference and mapping tool for researchers and practitioners of Artificial Intelligence. In addition, as researchers continue developing novel ways to combine Reinforcement Learning with Large Language Models, the tree can be potentially expanded with new nodes.

By bridging the gap between RL and LLMs, this taxonomy can be a valuable resource for researchers who are experienced in one of the two domains and are looking to venture into the other, and vice versa, as well as for anyone who wishes to explore whether a framework combining RL and LLMs is a promising solution for the problem they are seeking to address. More importantly, the taxonomy guides researchers as they are shaping the requirements of their application, whether those are related to model performance, training efficiency, or responsible AI considerations. 

\input{Supp-Tables/table-mapping}

%% file: Supp-Tables/table-mapping.tex
\begin{table*} 
\caption{Mapping Studies to RL/LLM Taxonomy Tree Leaves} \label {tab:studies}
\begin{center}
\begin{tabular}{llll}
\hline
\multicolumn{3}{l}{\textbf{RL/LLM Taxonomy}}                                                                      & \textbf{Studies}                   \\ \hline
\multirow{10}{*}{RL4LLM} & \multirow{5}{*}{Fine-tuning}        & \multirow{3}{*}{With Human Feedback}    & \citet{ouyang2022training}        \\ \cline{4-4}
                         &                                     &                                         & \citet{bai2022training}     \\ \cline{4-4} 
                         &                                     &                                         & \citet{hu2023aligning}         \\ \cline{3-4} 
                         &                                     & \multirow{3}{*}{Without Human Feedback} & \citet{bai2022constitutional}     \\ \cline{4-4} 
                         &                                     &                                         & 
                         \citet{ramamurthy2023reinforcement}     \\ \cline{4-4} 
                         &                                     &                                         & \citet{ghalandari2022efficient} \\ \cline{2-4} 
                         & \multirow{4}{*}{Prompt}             & \multirow{4}{*}{-}                      & \citet{zhang2022tempera}        \\ \cline{4-4} 
                         & & & \citet{deng2022rlprompt}        \\ \cline{4-4}
                         &                                     &                                         & \citet{sun2023offline}      \\ \cline{4-4} 
                         &                                     &                                         & \citet{perez2022red}      \\ \hline
\multirow{11}{*}{LLM4RL} & \multirow{4}{*}{Reward}             & \multirow{4}{*}{-}                      & \citet{kwon2023reward}        \\ \cline{4-4} 
                         &                                     &                                         & \citet{xie2023text2reward}       \\ \cline{4-4} 
                         &                                     &                                         & \citet{ma2023eureka}      \\ \cline{4-4} 
                         &                                     &                                         & \citet{song2023self}      \\ \cline{2-4} 
                         & Goal                                & -                                       & \citet{quartey2023exploiting}     \\ \cline{2-4} 
                         & \multirow{6}{*}{Policy}             & \multirow{6}{*}{-}                      & \citet{du2023guiding}        \\ \cline{4-4} 
                         &                                     &                                         & \citet{reid2022wikipedia}     \\ \cline{4-4} 
                         &                                     &                                         & \citet{hu2023language}         \\ \cline{4-4} 
                         &                                     &                                         & \citet{zhang2023rladapter}         \\ \cline{4-4} 
                         &                                     &                                         & \citet{carta2023grounding}      \\ \hline
                         
\multirow{4}{*}{RL+LLM}  & \multirow{1}{*}{Without Language Feedback} & \multirow{1}{*}{-}                       & \citet{yuan2023plan4mc}      \\ \cline{2-4}                       
                         &  \multirow{3}{*}{With Language Feedback}            & \multirow{3}{*}{-}                                     &  \citet{ahn2022i}        \\ \cline{4-4} 
                          & & & \citet{huang2022inner}        \\ \cline{4-4} 
                          & & & \citet{dasgupta2023collaborating} \\ \hline
\end{tabular}
\end{center}

\end{table*}

%% file: SECTIONS/4_RL4LLM.tex
\section{\texttt{RL4LLM}: Using Reinforcement Learning to Enhance Large Language Models} \label{sec:rl4llm}

As explained in section \ref{sec:background-contributions}, Reinforcement Learning with Human Feedback is an integral part for the training of Large Language Models. Nevertheless, there is a wealth of studies where the synergy between RL and LLM extends beyond training of the LLM. Therefore, \texttt{RL4LLM} class includes a significant body of work where Reinforcement Learning is used to further refine an already trained LLM and improve its performance on NLP-related tasks. This improvement in performance is measured according to the goal of each research, with the most common goals being the following:
\begin{itemize}
\item Improved performance in downstream NLP tasks \citep{deng2022rlprompt,ramamurthy2023reinforcement,ghalandari2022efficient,zhang2022tempera,sun2023offline}. 
\item Alignment with intent, values, and goals of user \citep{ramamurthy2023reinforcement,ouyang2022training}.
\item Alignment with Responsible AI considerations \citep{perez2022red,bai2022training,bai2022constitutional}.
\item Reduction of data and resource requirements \citep{zhang2022tempera,sun2023offline}.
\end{itemize}

\texttt{RL4LLM} studies can be further divided into two major sub-categories: a) Studies that use knowledge from LLMs to build an RL model to fine-tune an LLM, or part of it, to perform a downstream NLP task and b) studies using RL to design prompts to query LLMs.  A summary of the natural langauge application of each framework is shown in Table \ref{tab:rl4llm-apps}.

\subsection{\texttt{RL4LLM-Fine-Tuning}} \label{subsec:finetuning}
This class includes studies where RL is used for directly fine-tuning an existing LLM to make it more aligned with specific goals by updating an enormous set of LLM parameters. The presence of human feedback for fine-tuning serves as the criterion for further branching the \texttt{RL4LLM}-Fine-tuning node of our taxonomy tree, resulting in two new subclasses: \texttt{RL4LLM} - Fine-tuning with human feedback (\ref{subsec:withfeedback} and \texttt{RL4LLM} - Fine-tuning without human feedback (\ref{subsec:withoutfeedback}.

\subsubsection{With human feedback} \label{subsec:withfeedback}
Human input can be critical when assessing the quality of the LLM output in terms of harmlessness. Preventing the generation of toxic and harmful content has been the focus of multiple studies even before Large Language Models. For example, \cite{stiennon2020learning} trained an RL agent to predict which summary of a given Reddit post is most likely to be preferred by a human. The authors used a supervised learning model as a reward function that selects a summary among multiple candidate summaries. The result of the summary selection is then used to fine-tune the RL agent using PPO. Authors found that optimizing the reward model resulted in better human-preferred summaries in comparison to using typical NLP evaluation metrics like ROUGE. In a similar manner, the idea of fine-tuning language models through RL agents extended naturally to the realm of LLMs.

Human feedback can generally be embedded in the fine-tuning framework through the construction of the training dataset for the policy and reward models: For training the policy model, humans demonstrate the target behavior of the LLM, while for the reward model, they rank the alternative outputs of the LLM based on how well they align to the intent of the framework. For a study to be classified as \texttt{RL4LLM-Fine-Tuning} with human feedback, it should include human feedback in the training dataset of at least the initial policy model or the reward model; else, it belongs to  \texttt{RL4LLM-Fine-tuning} without human feedback.

\textbf{\citet{ouyang2022training}} developed Instruct-GPT, an LLM capable of capturing and following the intent of the user without producing untruthful, toxic, or generally unhelpful content. Instruct-GPT consists of three steps. The first includes the training of the policy model as a supervised learning model. The training dataset is generated by collecting demonstrations of the desired model behavior. To generate each new data point, a prompt is sampled from a baseline prompt dataset and a human ``labeler'' demonstrates the desired behavior of the model. The dataset is then used to fine-tune GPT-3 \citep{brown2020language} with supervised learning. The second step is the training of the reward model. Like the policy model, the reward model is also a supervised learning model, but it is trained on comparison data. To generate the training dataset of the reward model, a prompt and a set of model outputs are sampled for each data point, and a human ``labeler'' assigns rankings to the outputs.  Finally, the third step is  GPT-3 fine-tuning, with the reward model embedded in an RL training framework. Experimental evaluation in public NLP datasets showed that Instruct-GPT demonstrates improved performance regarding truthfulness and harmlessness compared to its baseline model, with only minimal performance degradation, while also showing generalization capabilities to instructions outside the distribution present in the fine-tuning dataset.

\textbf{\citet{bai2022training}} also utilized preference modeling and RLHF to train helpful, honest, and harmless AI assistants. Like \cite{ouyang2022training}, they trained an initial policy by fine-tuning a pretrained LLM. First, a HHH (Helpful, Honest, and Harmless) Context-Distilled Language Model was used to build a base dataset. This dataset was used to train a preference model to generate, in turn, a new dataset, using rejection sampling. Finally, the initial policy and the preference model were combined in an RLHF framework for fine-tuning the AI agent. Extensive data collection was performed by crowd workers, who were interacting with the models in open-ended conversations. The human feedback data, along with the preference models and the resulting RL policies, were updated on a weekly basis in an online manner to improve the quality of both the datasets and the models themselves. As a result, the authors achieved the desired alignment between the language model and human preferences in almost all NLP evaluations, with friendly and deployable AI assistants, that also presented improved AI capabilities in NLP tasks, such as text summarization, even extending to specialized tasks, like python code generation.

\textbf{\citet{hu2023aligning}} propose an offline RLHF framework to align LLMs to human intent. Rather than the typical PPO architecture applied in RLHF settings, they fine-tune the LLM on pre-generated samples in an offline manner, within a framework consisting of four steps: First, the pre-trained language model is fine-tuned on human-labeled instruction data using a supervised learning method, resulting in a new model called SFT. Second, they train a human preference model (RM) to predict rewards, using binary loss or ranking loss functions. Third, they build a combined dataset consisting of both human-labeled data (used in training the SFT model in the first step) as well as model-generated data (generated by the SFT model using prompts from the user, the SFT dataset, and the RM dataset). Finally, the SFT model is fine-tuned on this combined dataset using offline RL. The authors implemented three offline RLHF algorithms, namely Maximum Likelihood Estimation (MLE) with Filtering \citep{solaiman2021process}, Reward-Weighted Regression \citep{peters2007reinforcement}, and Decision Transformer \citep{chen2021decision} \ref{LLM4RL_policy}, with specific data pre-processing methods and training loss function for every algorithm choice. The performance of the models was evaluated both by humans and by GPT-4\citep{openai2023gpt4}, with the Decision Transformer architecture outperforming both MLE with Filtering and Reward-Weighted Regression in terms of evaluation score. The Decision Transformer-based offline method was also shown to obtain comparable results to PPO, while also achieving faster convergence.

\subsubsection{Without human feedback} \label{subsec:withoutfeedback}
This class of methods is primarily focused on the development of responsible AI systems. Interestingly, the presence of a human in the loop is not required for ensuring helpfulness and harmlessness of robotic assistants. As a result, this subclass includes studies where human feedback is either completely omitted or provided by a capable AI system. 


In a variation of \cite{bai2022training}, \textbf{\citet{bai2022constitutional}} proposed ``Constitutional AI'', a framework to train AI assistants capable of handling objectionable queries without being evasive by using AI Feedback. The AI model is trained through self-improvement, while human feedback is restricted in providing a list of rules and principles. Constitutional AI consists of two phases, namely a supervised learning phase and a reinforcement learning phase. In the supervised learning phase, an initial helpful-only LLM assistant generates responses to red teaming prompts that are designed to typically elicit harmful and toxic responses. This phase is the AI analogue of a human demonstrating the desired behavior in \cite{ouyang2022training} and \cite{hu2023aligning}, or the use of the distilled model \citep{bai2022training}. The model is asked to evaluate the response it provided based on a constitutional principle and then revise its response based on this critique. Responses are repeatedly revised at each step of the process by randomly drawing principles from the constitution. For the RL stage, a preference model is trained to act as the reward function using a training dataset generated by  the trained model from the supervised learning stage: To generate a data point, the assistant is prompted again with a harmful prompt and is asked to select the best response from pair of responses, based on the constitutional principles. This process produces an AI-generated preference dataset from harmlessness, which is combined with a human feedback-generated dataset for helpfulness. The preference model is then trained based on this combined dataset and is used to fine-tune the supervised model from the first stage in an RL framework that follows the general principles of RLHF - with the difference that the feedback is provided by the AI, hence the term ``RLAIF’’. The resulting LLM responds to harmful queries by explaining its objections to them. This study is an example where alignment to human goals can be achieved with minimal human supervision.

More recently, \textbf{\citet{ramamurthy2023reinforcement}} examined whether RL is the best choice for aligning pre-trained LLMs to human preferences, compared to supervised learning techniques, given the challenges of training instability that RL algorithms might suffer from, as well as the lack of open-source libraries and benchmarks that are suitable for LLM fine-tuning. To address those issues, the authors released RL4LM, an open-source library built on HuggingFace, which enables generative models to be trained with a variety of on-policy RL methods, such as PPO, TRPO, and A2C. The library provides a variety of reward functions and evaluation metrics. In addition, the authors composed GRUE (General Reinforced-language Understanding Evaluation) benchmark, a set of seven language generation tasks which are supervised by reward functions that quantify metrics of human preference. Finally, they introduce NLPO (Natural Language Policy Optimization), an on-policy RL algorithm (also available in RL4LM) that dynamically learns task-specific constraints over the distribution of language to effectively reduce the combinatorial action space in language generation. The authors provided detailed results on a case-by-case basis to determine when RL is preferred over supervised learning as well as when NLPO is preferred to PPO. However, NLPO demonstrated overall greater stability and performance than other policy gradient methods, such as PPO, while RL techniques were shown to generally outperform their supervised learning counterparts in terms of aligning LMs to human preferences.

\textbf{\citet{ghalandari2022efficient}} used Reinforcement Learning to fine-tune an LLM for sentence compression while addressing the issue of inefficiency at inference time. In the specific task of this study, the goal is to summarize a sentence by extracting a subset of its tokens in their original order. Given a tokenized input sentence x, the output is a binary vector indicating whether the corresponding input token is included in the compressed sentence or not. To evaluate the output of the policy, a reward is calculated as the average of three metrics: fluency (for grammatically correct and well-written sentences), similarity to source (to preserve the meaning of the original sentence, measured using bi-encoder similarity, cross-encoder similarity, and cross-encoder NLI), and output length or compression ratio (imposing soft length control using Gaussian reward functions). The policy was initialized using DistilRoBERTa \citep{sanh2020distilbert}, a six-layer a transformer encoder model with a linear classification head, and the RL agent was trained through a Policy Gradient method. The model was shown to outperform unsupervised models (with no labelled examples) while also enabling fast inference with one-step sequence labeling at test time and allowing for configurable rewards to adapt to specific use cases.

\subsection{\texttt{RL4LLM-Prompt}} \label{subsec:RL4LLM-prompt}
Constructing a suitable prompt is the first step towards ensuring that an LLM will generate the desired output in terms of relevance, format, and ethical considerations. Indeed, careful prompt engineering is often sufficient for aligning the output of an LLM to human preferences without fine-tuning the weights of the LLM itself, a computationally intensive process which usually requires extensive data collection. Prompting concatenates the inputs with an additional piece of text that directs the LLM to produce the desired outputs. Most studies focus on tuning soft prompts (e.g., embeddings), which are difficult to interpret and non-transferable across different LLMs \citep{deng2022rlprompt}.  On the other hand, discrete prompts, which consist of concrete tokens from vocabulary, are hard to optimize efficiently, due to their discrete nature and the difficulty of efficient space exploration. To address this limitation, this class of studies utilize RL for discrete prompt optimization, with the goal to enhance the performance of the LLM on diverse tasks, often requiring just a few training instances. 

Two recent studies in this class are TEMPERA and RLPROMPT, both of which use Roberta-large as the background LLM. Proposed by \textbf{\citet{zhang2022tempera}}, TEMPERA (TEst-tiMe Prompt Editing using Reinforcement leArning) is a framework for automating the design of optimal prompts at test time. Prompt optimization is formulated as an RL problem with the goal to incorporate human knowledge and thus design prompts that are interpretable and can be adapted to different queries. The RL agent performs different editing techniques at test time to construct query-dependent prompts efficiently. The action space allows the RL agent to edit instructions, in-context examples and verbalizers as needed, while the score differences between successive prompts before and after editing are used as reward. Unlike previous methods, TEMPERA makes use of prior human knowledge and provides interpretability; also, compared to approaches like prompt tweaking, AutoPrompt, and RLPrompt, it also significantly improves performance on tasks like sentiment analysis, subject classification, natural language inference, etc. 

On the other hand, RLPROMPT by \textbf{\citet{deng2022rlprompt} }is an optimization approach where a policy network is trained to generate desired prompts. The experimental results show that the policy is transferable across different LMs, which allows learning cheaply from smaller models and infer for larger, more powerful models. The authors also noted that optimized prompts were often grammatical “gibberish”, which indicates that high-quality LM prompting does not necessarily follow human language patterns. However, RLPROMPT treats the LLM as a black box model with only access to the generated output whereas TEMPERA assumes it to have access to the embedding vectors. The policy models are similar with GPT encoder but the action space is very different, since TEMPERA uses only discrete actions, whereas RLPROMPT treats the entire vocabulary as possible actions. Finally, the performance of TEMPERA was benchmarked for text classification tasks, whereas RLPROMPT was applied to text generation.

More recently, \textbf{\citet{sun2023offline}} proposed Prompt-OIRL, a framework that uses offline reinforcement learning to achieve cost-efficient and context-aware prompt design. The framework utilizes readily available offline datasets generated through expert evaluation of previously crafted prompts. First, the authors apply inverse-RL to learn a proxy reward model that can perform query-dependent offline prompt evaluations. Then, they use this model as an offline evaluator to perform query-dependent prompt optimization. Contrary to \cite{deng2022rlprompt}, who perform task-agnostic prompt optimization, Prompt-OIRL performs query-dependent prompt evaluation, similarly to \cite{zhang2022tempera}. The dependence of prompt evaluation on the query allows for context awareness, which helps the prompt evaluator predict what prompting techniques (e.g., Chain of Thought) are most likely to obtain correct answer given a specific prompt (e.g., an arithmetic question). The design of the proxy reward function allows for offline query-dependent evaluation thus achieving both context awareness and lower costs, and the framework was evaluated across four LLMs and three arithmetic datasets.

A side-by-side comparison of the three methods is shown in Table \ref{tab:prompt}.
\input{Supp-Tables/table-prompt}

Another study where Reinforcement Learning with AI feedback is used to ensure harmlessness of AI assistants – this time by using RL for prompt design - is the study of \textbf{\citet{perez2022red}}, who used a Language Model to generate test questions (``red teaming'') that aim to elicit harmful responses from the target LM. Then, a classifier that is trained to detect offensive content is used to evaluate the target LM’s replies to those questions. Contrary to studies in previous sections, this study uses RL to train the red-teaming LLM, instead of fine-tuning or prompting the target LLM. More precisely, starting from a pretrained LM for red teaming, the authors perform a first pass of fine-tuning using supervised learning. Then, they use RL to train the red-teaming LLM in a synchronous advantage actor-critic (A2C) framework with the objective of maximizing the expected harmfulness. The reward function is a linear combination of the A2C loss and the K-L divergence penalty between the target policy and the distribution of the initialization over the next tokens. The authors performed red teaming for a variety of harmful behaviors, including offensive language, data leakage, personal contact information generation, and distributional bias of the output text. LM-based red teaming was shown to be a promising tool for the timely identification of potentially harmful behavior, with RL-based red teaming being the most effective at eliciting offensive replies compared to the other methods, which included zero-shot and stochastic few-shot generation, and supervised learning.

\input{Supp-Tables/table-rl4llm-aps}

%% file: Supp-Tables/table-prompt.tex
\begin{table*} 
\caption{RL4LLM-Prompt Studies} \label {tab:prompt}
\begin{center}
\renewcommand{\arraystretch}{1.5}
\begin {tabularx}{0.8\textwidth}{XXXX} \hline
\textbf{Dimensions}&	\textbf{TEMPERA} \cite{zhang2022tempera} &	\textbf{RLPROMPT} \citep{deng2022rlprompt} &  \textbf{Prompt-OIRL} \citep{sun2023offline} \\ \hline
LM Model&	Roberta-large &		Roberta-large & GPT-3.5, TigerBot-13B-chat, Llama2-7B-chat \\ \hline
Assumptions on LLM &	Not a black-box model, access to the hidden states	&	Black box model with no access to gradient & Black box model with no access to gradient \\ \hline
Policy Model&	GPT encoder with attention over all possible actions &	Frozen distilGPT-2 (82M) with one MLP layer on top (tunable) & Choose one of the K prompts, (CoT, ToT, etc.) \\ \hline
Action Space &	Discrete actions, e.g., swap, add or delete tokens	&	Tokens from the policy models vocabulary & Apply a prompt to an input query \\ \hline
RL Algorithm	&	PPO & Soft Q-learning & Offline Inverse RL \\ \hline
Applications &	Only Text Classification &	Text Classification and Text generation & Arithmetic reasoning\\ \hline

\end{tabularx}
\end{center}
\end{table*}

%% file: Supp-Tables/table-rl4llm-aps.tex
\begin{table*} 
\caption{RL4LLM Natural Language Processing Application} \label {tab:rl4llm-apps}
\begin{center}
\renewcommand{\arraystretch}{2}
\begin{tabularx}{\textwidth}{lX}
\hline
\textbf{LLM4RL Study}   & \textbf{Application} \\ \hline
\citet{ouyang2022training} & Generation, Open and Closed Question-Answering, Brainstorming, Chat, Rewrite, Summarization, Classification, Extraction  \\ \hline
\citet{bai2022training} &  Dialogue (AI Assistant) \\ \hline
\citet{hu2023aligning} &  Question-Answering \\ \hline
\citet{bai2022constitutional} &  AI assistant \\ \hline
\citet{ramamurthy2023reinforcement} & GRUE Benchmark tasks: Text Continuation, Generative
Commonsense, Summarization, Data to Text, Machine Translation, Question-Answering, Chitchat Dialogue   \\ \hline
\citet{ghalandari2022efficient} &  Sentence Compression (Sentence summarization, Text simplification, Headline generation)  \\ \hline
\citet{zhang2022tempera} &  Sentiment Analysis, Topic Classification, Natural Language Inference, Reading Comprehension \\ \hline
\citet{deng2022rlprompt} &  Few-Shot Text Classification, Unsupervised Text Style Transfer \\ \hline
\citet{sun2023offline} & Arithmetic reasoning (MultiArith \cite{roy2016solving}, GSM8K \cite{cobbe2021training}, SVAM \cite{patel2021nlp})\\ \hline
\citet{perez2022red} & Question-Answering \\ \hline

\end{tabularx}
\end{center}

\end{table*}

%% file: SECTIONS/5_LLM4RL.tex
\section{LLM4RL: Enhancing Reinforcement Learning Agents through Large Language Models} \label{sec:llm4rl}

The \texttt{LLM4RL} class covers studies where an LLM acts as a component of an RL training pipeline. Contrary to \texttt{RL4LLM} studies (section \ref{sec:rl4llm}, where the end goal is an NLP task, the RL agent in this category is trained for  tasks which are generally not related to natural language. Overall, the motivation behind studies in the \texttt{LLM4RL} category is twofold, as shown on Table \ref{tab:llm4rl-goals}.
\begin{enumerate}
\item Improved performance of the RL Agent: In \texttt{LLM4RL} frameworks, improving the performance of the agent requires alignment with human intent or feedback \citep{hu2023language,song2023self,ma2023eureka,kwon2023reward,xie2023text2reward}, grounding of the agent to its environment \citep{xie2023text2reward,carta2023grounding} or learning learning complex tasks \citep{dasgupta2023collaborating,ma2023eureka}
\item Efficient training of the RL Agent: Training an RL agent can be computationally intensive, requiring not only significant computational resources, but also large amounts of data. Even with those prerequisites available, RL training might still suffer due to inefficient sampling, especially for complex, long-term tasks. Therefore, \texttt{LLM4RL} frameworks also focus on improving training efficiency – and therefore ensuring satisfying execution of the target tasks at test time – by facilitating exploration \citep{quartey2023exploiting,du2023guiding}, policy transfer of trained models \citep{reid2022wikipedia} and effective planning for reduced data requirements \citep{dasgupta2023collaborating}.
\end{enumerate}

The LLM replaces or assists in different ways one of the fundamental components of the reinforcement learning agent - namely, the reward function, the training goal, or the policy function. 
Using the corresponding component in each case as a criterion, we further break down the \texttt{LLM4RL} class in three sub-categories, where the LLM is used for a) determining the reward function (\texttt{LLM4RL-Reward}), b) expressing internal goals (\texttt{LLM4RL-Goal}), and c) pretraining, representing, or updating the policy function (\texttt{LLM4RL-Policy}).

\subsection{\texttt{LLM4RL-Reward}} \label{subsec:reward}
As noted by \cite{suttonbarto}, ``the use of a reward signal to formalize the idea of a goal is one of the most distinctive features of reinforcement learning''. The reward signal received through the interaction with the environment is critical for training an agent to achieve the desired behavior. Until recently, the bulk of RL research treated the reward function as given and focused on the training algorithms themselves \citep{eschmann2021}. Designing the reward function of an RL training framework is straightforward when direct knowledge of the problem is available, as when solving a well-defined problem or earning a high score in a well-defined game. Common examples in this category are Atari games, which are frequently utilized as sandbox environments for testing various aspects of RL training, or games where the agent receives a positive reward if they win, and negative reward otherwise.  However, there exists a significant number of applications where it is difficult to directly translate the desired behavior into rewards signals, especially for long and complex tasks or when the agent can discover unexpected ways, and potentially dangerous, ways to generate reward from the environment.

When the agent must learn to perform a long and possibly complex task, where designing a proper reward function not straightforward, and where human feedback is critical, it is common to rely on expert demonstrations or interactive modification of the reward signal. Expert demonstrations generally utilize on a technique known as Inverse Reinforcement Learning to infer the reward function by observing the desired behavior \citep{abbeel2004apprenticeship}, with a reward designer observing the agent’s performance and tweaking the reward signal during a trial-and-error process to asjust the agent's behavior accordingly.

Motivated by the direct involvement of humans in this interactive loop, combined with the ability of LLMs to learn in-context from few or even zero examples \citep{brown2020language}, researchers are exploring ways to bridge the gap between human preference and agent behavior in cases where the explicit quantification of rewards is difficult or time-consuming. In this context, the LLM is used either for reward shaping, i.e., i.e., guiding the learning agent with additional rewards to preserve policy optimality, or as a proxy reward function. Prior to the LLM era, \cite{goyal2019using} used language for reward shaping: They trained a model to predict if the actions in a trajectory match some specific language description and used the output to generate intermediate RL rewards. Similarly, \cite{cao2023temporal} extended the underlying Markov Decision Process by including a natural language instruction; the authors first generated instructions and obtained the corresponding word embeddings using BERT, and then trained an alignment model that maps action trajectories to their corresponding instructions. They found that augmenting the default reward of an Atari environment with the language-based reward significantly improves the performance of the agent.

The first study utilizing LLMs for RL agent reward design is the one by \textbf{\citet{kwon2023reward}}, who evaluated whether LLMs can produce reward signals that are consistent with the user behavior using GPT-3 \citep{gpt3} as a proxy reward function in an RL framework.  At the beginning of training, the user specifies the desired behavior using a prompt with explanation and an example of the desired behavior, while during training, the LLM evaluates the agent’s behavior against the behavior described in the prompt and generates a reward signal accordingly, which the RL agent uses to update its behavior. In more detail, the proposed framework is as follows: The LLM is provided with a task description, a user’s description of the objective, a string-formatted episode outcome, and a question asking if the outcome satisfies the objective. The LLM’s response to the question is used as reward signal for the agent. Based on this reward signal, the agent updates their weights and generates a new episode, the outcome of which is parsed back into a string, and the episode continues. To evaluate their framework, the authors compared it to three baseline cases: a) a few-shot baseline, where a supervised learning model is trained to predict reward signals using the same examples given to the LLM, b) a zero-shot baseline, where the LLM is prompted without the user’s description of the objective, and c) a baseline where the agents are trained with ground truth reward functions. Experimental results showed that the proposed RL training framework - which is agnostic to the RL algorithm used - can achieve user objective-aligned behavior, as measured both with automated metrics and with human users. In addition, the performance of the agents is shown to outperform agents trained with reward functions learned via supervised learning, even when no examples were provided - as long as the objective is well-defined - or when the tasks were complicated.

Grounding a robotic agent to the environment and achieving the desirable behavior as directed through human feedback is the focus of TEXT2REWARD framework by \textbf{\citet{xie2023text2reward}}. TEXT2REWARD allows for the generation and continuous improvement of python code that expresses dense rewards functions for robotic agents performing manipulation tasks. The framework is composed of three stages: Abstraction, Instruction, and Feedback. In the Abstraction stage, an expert provides an abstract representation of the robot’s environment using Python classes. In the Instruction stage, a user provides a natural language description of the goal to be achieved by the robot (e.g., ``push the chair to the marked position''). Finally, in the Feedback phase, the user summarizes their preferences or the failure mode of the robot’s action. This summary is then used to update the reward function and retrain the RL agent. The authors evaluated their framework on two robotic manipulation benchmarks - MANISKILL2 \citep{gu2023maniskill2} and METAWORLD \citep{yu2021metaworld} - and two MUJOCO locomotion environments \citep{brockman2016openai}. For manipulation tasks, the experiments demonstrated comparable results to human oracle-designed rewards in terms of performance and convergence speed, with the performance improvement verified through few-shot examples. For locomotion tasks, the agent was able to successfully learn six new behaviors (move forward, front flip, back flip, etc.) with a rate of success ranging from 94\% to 100\% for each task. Finally, error analysis revealed that the generated code was correct at 90\% of the time, with most common errors originating from wrong use of class attributed (wrong use or hallucination of non-existent attributes), syntax errors or shape mismatch, or wrong imports. The success of TEXT2REWARD was largely owed to the use of human feedback to resolve ambiguity through providing clear, language-based instructions to correct the behavior of the robot. Aside from grounding and alignment to human preferences, TEXT2REWARD has the advantage of generating of highly interpretable functions, while requiring any data for reward training.

The EUREKA framework by \textbf{\citet{ma2023eureka}} is another example of reward design through direct python code generation. EUREKA consists of three fundamental components: the use of environment as context, evolutionary search, and reward reflection. The environment code, excluding the part of it that defines the reward, is directly provided to the LLM, which in turn extracts its semantics to compose a suitable reward function for the target task.  The LLM outputs the reward code, following some general formatting instructions. Evolutionary computation is used to overcome sub-optimal or non-executable code by sequentially generating improved reward functions using the concepts of reward mutation, which modifies previous solutions based on textual feedback, and random restarts, to escape local optima. Finally, reward reflection acts as a supplement to the numeric value of the reward expressed through the fitness function by explaining why a candidate reward function works or does not work and assigning credit accordingly. The framework follows a PPO architecture and is tested on a variety of benchmarking environments (e.g., Cartpole and BallBalance), where it achieves a higher success rate compared to human-specified rewards. The use of evolutionary computation is shown to be necessary for continuous increase in the success rate of the framework over time, while also allowing for the generation of more diverse and often counter-intuitive rewards that outperform human-designed rewards, particularly for difficult tasks. The authors also implemented a curriculum learning \citep{bengio2009} approach to teach a Shadow Hand to rotate a pen according to a set of pre-defined spinning patterns, thus demonstrating the capability of the framework to execute complex, low-level tasks. By successfully handling task complexity and allowing for the discovery of unexpected high-performing policies, EUREKA successfully deals with two key reasons that inhibit the translation of the desired agent behavior to rewards that were identified in subsection \ref{subsec:reward}. Finally, similarly to the \texttt{LLM4RL}-Reward studies discussed in \ref{subsec:reward}, a key benefit of EUREKA is the alignment of rewards to human preferences by incorporating human knowledge about the state through appropriate initialization of the reward. 

Similarly, \textbf{\citet{song2023self}} proposed a three-step, self-refined LLM framework for generating reward functions to help robotic agents achieve specific goals. The first step is the initial design of the reward function based on the natural language input provided by the user. The input includes a description of the environment, a description of the task in hand, broken down into goals, a description of the observable state (such as position and velocity of robot), and a list of rules that should be followed when designing the reward function (for example, restricting the dependence of the reward exclusively on known quantities). In the second step, the initial reward function is applied, the robot acts, and its behavior is evaluated. In the evaluation step, the user collects their observations around the training process and convergence, the objective metrics, the success rate   of the tasks, and makes an overall assessment (``good'' or ``bad'') of the robotic agent’s performance. Finally, in the self-refinement step, the feedback of the user is embedded in a feedback prompt, which is then used by the LLM to generate the next reward signal. The authors evaluated the performance of the framework for nine different continuous control tasks across three different robotic system systems and achieved a success rate between 93\% and 100\% for all tasks, outperforming the corresponding manual reward setting in almost all of them.

\subsection{\texttt{LLM4RL-Goal}}

Goal setting is a key element in intrinsically motivated reinforcement learning \citep{NIPS2004_4be5a36c}, which addresses key challenges of the traditional Deep RL setting, including the difficulty to abstract actions and to explore the environment \citep{aubret2019survey}. Contrary to traditional RL, where the training of the agent relies exclusively on external rewards from the environment, intrinsic RL builds on the psychology concept of intrinsic motivation and developmental learning, which is inspired by babies developing their skills by exploring the environment \citep{aubret2019survey}. In a similar manner, Intrinsic Motivation in RL allows agents to learn reusable, general skills that are applicable to various tasks over their lifetime. With right internal goals, agents can autonomously explore open-ended environments and build useful skills at a pretraining phase. Intrinsic RL is therefore particularly useful when the design of a reward function is not straightforward. However, the benefits of intrinsic RL are limited when the environment complexity and size increases significantly, since there is no guarantee that the skills the agent builds throughout its lifetime will be useful to perform any downstream tasks at all.

To address this shortcoming, \textbf{\citet{du2023guiding}} proposed a new, intrinsically motivated RL method called ELLM (Exploring with LLMs). The authors build on the observation that, when faced with new tasks, humans do not uniformly (and therefore, blindly) explore the outcome spaces of their actions, but rely on their physical and social common sense to prioritize the exploration of plausibly useful behaviors, that have the reasonably highest likelihood to succeed, such as using a key to open a door. ELLM leverages knowledge from text corpora to capture useful semantic information and thus enable structured exploration of task-agnostic (i.e., pretraining) environments by allowing the agent to use this knowledge to reason about the usefulness of new begaviors. In particular, the authors used a pre-trained LLM, GPT-2 \citep{Radford2019LanguageMA} to suggest goals during exploration. In each step, the LLM is prompted with a list of the agent’s available actions, along with a text description of the current observation, and suggests a goal, which should be diverse, common-sense sensitive, and context sensitive. After the acgent takes an action and transitions in the environment, the goal-conditioned rewards are computed based on the semantic similarity between the LLM-generated goal and the description of the agent’s transition. The exploration of RL training is therefore controlled and enhanced without the need for explicit human intervention. ELLM was shown capable of producing context-sensitive, common-sensical and diverse goals, boosting pretraining exploration performance, as well as performance on downstream tasks.

TaskExplore by \textbf{\citet{quartey2023exploiting}} is another framework that uses LLMs to facilitate RL training by generating intermediate tasks. The goal of TaskExplore is to maximize the use of previous experiences collected by a robotic agent during their training in a simulated home environment by generating and learning useful auxiliary tasks while solving a larger target task. The process starts by converting a task specification to a graph using Linear Temporal Logic. Then, the graph is traversed to create an abstract task template where object propositions are replaced with their context-aware embedding representations, which have been generated by an LLM. The abstract task template is subsequently used to generate auxiliary tasks for each proposition node by swapping relevant objects from the environment and selecting objects with the most similar embeddings with the template embedding node under consideration. The training process consists of both online RL (where a policy for the given task is learned by following an epsilon-greedy behavioral policy) and offline RL (where all Q-value functions, including those of subtasks, are updated). Interestingly, the authors found that learning the auxiliary tasks does not adversely affect the performance of learning the main task. In addition, the auxiliary tasks learned through TaskExplore performed better compared to the same tasks learned through a random behavior policy. Finally, the curriculum built on auxiliary tasks developed through TaskExplore outperformed the corresponding  curriculum developed by randomly sampled tasks.

\subsection{\texttt{LLM4RL-Policy}} \label{subsec:LM4RLpolicy}
In this subclass, a large language model directly assists the policy of an RL agent by generating trajectories for pretraining \citep{reid2022wikipedia}, creating a policy prior \citep{hu2023language}, acting as a planner in a Planner-Actor-Reporter scheme \citep{dasgupta2023collaborating}, by directly representing the policy  \citep{carta2023grounding}, or by combining an adapter model to fine-tune the prompt of another LLM that generates instructions \citep{zhang2023rladapter}.

\subsubsection{Trajectory Generation for Pretraining}
The power of Reinforcement Learning largely lies in the ability of the agent to learn through interacting with its environment. However, sometimes the interaction with the environment is impractical, either because it is expensive (e.g., in robotics), or dangerous (as in healthcare or autonomous vehicles) \citep {levine2020offline}. However, as explained in the \texttt{LLM4RL-Reward} subsection (\ref{subsec:reward}, even when such interaction is possible and assuming available training data and computational resources, training a model from scratch might still face slow convergence. Therefore, pretraining with previously collected data can benefit the agent, particularly in complex domains where large datasets are needed for scalable generalization. In such cases, Offline Reinforcement Learning is commonly used.  Offline RL handles the control problem as a sequence modeling problem \citep{chen2019hardware,chen2021hardware,furuta2022generalized} and uses supervised learning to fit a dataset consisting of state-action-reward trajectories.

By framing offline RL as a sequence modeling problem, \textbf{\citet{reid2022wikipedia}} investigate whether LLMs can be successfully transferred to other domains when fine-tuned on offline RL tasks that have no relation to language. The authors modeled trajectories autoregressively, representing each trajectory $t$ as a sequence of the form $t = (\hat{R}_1, \allowbreak s_1, \allowbreak \alpha_1, \allowbreak\hat{R}_2, s_2, \alpha_2, \allowbreak ...,\hat{R}_N, s_N, \allowbreak \alpha_N)$, with $\hat{r}_i$, $s_i$, and $a_i$ representing the state, action, and cumulative reward, respectively, at time step $t$. The final objective of the supervised learning was a weighted sum of three loss functions: The primary loss function is a classic Mean Squared Error loss. The second loss function is a cosine similarity loss that quantifies the similarity between language representations and offline RL input representations, with the goal to make the input language embeddings as similar as possible to their language counterparts. Finally, the third loss function represents the negative log=likelihood-based language modeling objective to allow for joint training of language modeling and trajectory modeling. The authors used pre-trained models GPT2-small and ChibiT, a model that was pretrained on Wikitext-103 dataset \citep{merity2016pointer}. As shown by experiments in  four Atari tasks and three OpenAI Gym tasks, pretraining with language datasets can successfully fine-tune offline RL tasks and, most importantly, by achieving significant gains in terms of both convergence speed and total reward received, compared to baseline offline RL  including Decision Transformer \citep{chen2021decision}, CQL \citep{kumar2020conservative}, TD3+BC \citep{fujimoto2021minimalist}, BRAC \citep{wu2019behavior}, and AWR baselines \citep{peng2019advantageweighted}.


\subsubsection{Creating a Policy Prior}
In subsection \ref{subsec:reward}, we reviewed studies where human-agent alignment is achieved through reward design. An alternative way to train RL agents that behave according to human preferences is the creation of a policy prior that is aligned with human preferences and reasoning. \textbf{\citet{hu2023language}} propose Instruct-RL, a framework for human-AI coordination where the human uses high-level natural language instructions to specify to the AI agent the type of behavior they expect. The human natural language instruction is then passed to pre-trained LLMs, which produce a prior policy. To construct the prior, the LLM is supplied with the initial human instruction as well as a language prompt, which provides a language description of the observations of the environment while the policy is being followed. In practice, the LLM uses a softmax function to predict the probability of possible actions given the observed state and the human instruction. Finally, the policy prior is used as a reference policy during the training of the RL agent to regularize the training objective, with the regularization technique varying according to the RL training algorithm, by augmenting the epsilon-greedy method in Q-learning and adding a KL penalty to the training objective in PPO. The experimental results based on both the performance of the algorithms in benchmarking tasks (Hanabi and a negotiating game) and the human evaluation demonstrated that instructRL successfully incorporated human preference to produce high-performance policies, even when the prior policies were imperfect and generated with simple prompts. However, the authors highlighted the need for fine-tuning to improve test-time performance, since adding the LLM priors to trained agents was shown to add no meaningful improvement to the policy.

\subsubsection{LLM Being the Policy} \label{LLM4RL_policy}

While LLMs possess general reasoning capabilities, they are not trained to solve environment specific problems during their training, and thus cannot influence or explore the specific environment where a task needs to be accomplished. \textbf{\citet{carta2023grounding}} proposed a framework to overcome the lack of alignment between the general statistical knowledge of the LLM and the environment by functionally grounding LLMs and using them directly as the policy to be  updated. Using a simple grid-world based text environment (Baby AI), the authors formulated a goal-augmented, partially observable MDP where, given a prompt p, the LLM outputs a probability distribution over the possible actions and an action is sampled according to this distribution. The authors used Proximal Policy Optimization to train the agent and used Flan-T5 780 M as the policy LLM and observed 80\% success rate after 250 K training steps, which is a significant improvement compared to previous models. The authors also considered genitalization over new objects, where a drop in performance was observed, with the model still outperforming the benchmark. The biggest drop is observed when testing against generalization to new tasks or a different language.

\subsubsection{Planner}

\textbf{\citet{dasgupta2023collaborating}} combined the reasoning capabilities of the LLM with the specialization and the knowledge of the environment of an RL agent in a Planner-Actor-Reporter scheme. This study belongs to the \texttt{RL + LLM} cateogry, and will therefore be analyzed in section \ref{sec:llm4rl}. However, the authors also designed a variation of the main framework where the Planner is embedded in the training loop of the Reporter with the goal to increase its truthfulness, so that it reports accurate information to the planner. The Reporter uses the reward received at the end of the episode to update their reporting policy such that it eventually learns to only report helpful, i.e. truthful and relevant, information back to the planner. The reader shall refer to section \ref{sec:rlplusllm} for more details on this publication.

\subsubsection{Using an adapter model}
\textbf{\citet{zhang2023rladapter}} developed the RLAdapter Framework, a complex system that, apart from the RL agent and the LLM, includes additionally an Adapter model to improve the connection between the RL agent and the LLM without requiring the costly and often impossible fine-tuning of the base LLM. The Adapter itself is also an LLM - in particular, a 4-bit quantized version of the LLaMA2-7B model, which is fine-tuned with feedback from the RL agent and the LLM. Embedding the adapter model in the RL training loop is shown to result in more meaningful guidance overall, by enhancing both the LLM's comprehension of downstream tasks as well as the agent’s understanding capability and effective learning of difficult tasks. The key metric that aids this closed-loop feedback is the understanding score, which quantifies the semantic similarity between the agent’s recent actions and the sub-goals suggested by the LLM as measured by the cosine similarity between the embeddings of the LLM-provided sub-goals and the episode trajectory. The prompt of the Adapter includes a description pf the player's observations, past actions, past sub-goals, and the understanding score. The Adapter generates a prompt for the base LLM, containing a natural language summary of the player's past observations, actions, and understanding. In turn, the base LLM generates updated instructions for the RL agent, which, as usual, takes an action, receives the response from the environment, and updates its policy. The understanding score is then calculated and is used to fine-tune the adapter model, which then goes on to generate a new prompt. The performance of RLAdapter with GPT-3.5 \citep{gpt35} as baseline was compared to baseline models, including ELLM by \cite{du2023guiding}. RLAdapter was shown to outperform all baselines for 1 million steps, apart from ELLM; however, for 5 million steps, the performance of RLAdapter with GPT-4 \citep{openai2023gpt4} exceeded that of ELLM as well, and RLAdapter with GPT-3.5 \citep{gpt35} matched SPRING by \cite{wu2023spring} in terms of performance.

The main novelty of RLAdapter lies in fine-tuning the lightweight, cheaper-to-update adapter model, while only updating the prompt of the base LLM. Despite the LLM prompt being updated through this feedback loop, we do not classify this study as \texttt{RL4LLM}, since RL is not used to improve the performance of a downstream language task.

\input{Supp-Tables/table-LLM4RL}
\input{Supp-Tables/table-llm4rl-apps}

\input{Supp-Tables/table-llm4rl-base}

%% file: Supp-Tables/table-llm4rl.tex
\begin{table*} 
\caption{Breakdown Of LLM4RL Synergy Goals Per Subclass} \label {tab:llm4rl-goals}
\begin{center}
\renewcommand{\arraystretch}{1.5}
\begin{tabularx}{\textwidth}{XXXXXX}
\hline
 Study & \multicolumn{3}{l}{\textbf{Test-Time Performance}}   & \multicolumn{2}{l}{\textbf{Training Efficiency}} \\ \hline
 & \textit{Alignment with human intent} & \textit{Grounding} & \textit{Learning complex tasks} & \textit{Improving exploration} & \textit{Policy Reuse } \\ \hline

\citet{kwon2023reward}  & \checkmark & & &  \\ \hline
\citet{xie2023text2reward} & \checkmark & \checkmark & &  \\ \hline
\citet{ma2023eureka}  & \checkmark & & \checkmark &  \\ \hline
\citet{song2023self}  & \checkmark & & &  \\ \hline
\citet{du2023guiding}  & & & & \checkmark  \\ \hline
\citet{quartey2023exploiting}  & & & \checkmark &  \\ \hline
\citet{reid2022wikipedia}  & & & & & \checkmark \\ \hline
\citet{hu2023language}  & \checkmark & & &  \\ \hline
\citet{carta2023grounding}  & & \checkmark & &  \\ \hline
\citet{zhang2023rladapter}  & & \checkmark & &  \\ \hline
\end{tabularx}
\end{center}
\end{table*}


%% file: Supp-Tables/table-llm4rl-apps.tex

\begin{table*} 
\caption{LLM4RL Environment and Applications} \label {tab:llm4rl-apps}
\begin{center}
\scalebox{0.8}{
\begin{tabularx}{\textwidth}{lX}
\hline
\textbf{LLM4RL Study}   & \textbf{Environment and Application} \\ \hline
 \citet{kwon2023reward} & Ultimatum Game, 2-Player Matrix Games, DEALORNODEAL negotiation task \citep{lewis2017deal}\\ \hline
 
\citet{xie2023text2reward} & 17 manipulation tasks: Two robotic manipulation benchmarks - MANISKILL2 \citep{gu2023maniskill2} and METAWORLD \citep{yu2021metaworld} and two locomotion environments of MUJOCO \citep{brockman2016openai}.
1)	METAWORLD: Benchmark for Multi-task Robotics Learning and Preference-based Reinforcement Learning. Robot arm for tabletop tasks: Drawer Open/Close, Window Open/Close, Button Press, Sweep, Door Unlock/Close, Handle Press/Press Slide .
2)	MANISKILL2: object manipulation tasks in environments with realistic physical simulations. Tasks: Lift/Pick/Stack Cube, Turn Faucet, Open Cabinet Door/Drawer, Push Chair
3)	Gym MuJoCo: Hopper (Move Forward, Front Flip, Back Flip), Ant (Move Forward, Wave Leg, Lie Down).
4)	Real Robot. Manipulation tasks - Pick-and-place, assembly, articulated object manipulation with revolute or sliding joint, and mobile manipulation, \\ \hline
\citet{ma2023eureka} & 
For Isaac Gym Environments: Cartpole, Quadcopter, FrankaCabinet, Anymal, BallBalance, Ant, AllegroHand, Humanoid, ShadowHand, Over, DoorCloseInward, DoorCloseOutward, DoorOpenInward, DoorOpenOutward, Scissors, SwingCup, Switch, Kettle, LiftUnderarm, Pen, BottleCap, CatchAbreast, CatchOver2Underarm, CatchUnderarm, ReOrientation, GraspAndPlace, BlockStack, PushBlock, TwoCatchUnderarm. Pen spinning as a complicated dexterous task.
\\ \hline
\citet{song2023self} & Three robotic systems: 1) Robotic Manipulator for ball catching, ball balancing, and ball pushing (Franka Emika Panda Emika  by \cite{frankaemika}), 2) Quadruped robot for velocity tracking, running, and walking to target (Anymal AnyRobotics by \cite{anyrobotics2023}, 3) Quadcopter for hovering, flying through a wind field, and velocity tracking (Crazyflie BitCraze by \cite{bitcraze2023}). \\ \hline
\citet{du2023guiding} & 1) Crafter game environment (2-D version of Minecraft), modified to a) replace the general ``Do'' command with more specific commands and b) increase damage against enemies and reduce the amount of wood needed to craft a table. 2) Housekeep robotic simulator by \cite{kant2022housekeep}, where an agent cleans up a house by rearranging objects. \\ \hline
\citet{quartey2023exploiting} & HomeGrid. Used a food preparation task (maps to visiting the right squares in the right order). \\ \hline
\citet{reid2022wikipedia} & Two multi-agent coordination games: Say-Select (a cooperative game where two players are collecting rewards by selecting from a set of balls, each of which is mapped to a reward value that is only known to one of the players) and Hanabi \citep{bard2020hanabi}, a cooperative card game.\\ \hline
 \citet{hu2023language} & Multi-agent coordination games: Say-Select and Hanabi. \citep{bard2020hanabi} \\ \hline
\citet{carta2023grounding} & New environment introduced: BabyAI-Text (Adaptation of BabyAI \citep{chevalierboisvert2019babyai} to be text-only). Minigrid environment where an agent navigates and interacts with objects through 6 commands: turn left, turn right, go forward, pick up, drop, toggle.  \\ \hline
\citet{zhang2023rladapter} & Crafter Game with 22 different tasks (e.g., collecting resources, crafting items, defeating monsters) \\ \hline

\end{tabularx}
}
\end{center}

\end{table*}

%% file: Supp-Tables/table-llm4rl-base.tex
\begin{table*} 
\caption{LLM4RL Base Algorithms and RL Architecture} \label {tab:rl4llmbase}
\begin{center}
\renewcommand{\arraystretch}{1.5}
\begin{tabularx}{\textwidth}{lXX}
\hline
\textbf{LLM4RL Study}   & \textbf{Base LLM} & \textbf{RL Architecture} \\ \hline
\citet{kwon2023reward} & GPT-3 \citep{gpt3} & DQN \citep{mnih2013playing} \\ \hline
\citet{xie2023text2reward} & GPT-4 \citep{openai2023gpt4} & PPO \citep{schulman2017proximal} \citep{schulman2017proximal}, SAC \citep{haarnoja2018soft} \\ \hline
\citet{ma2023eureka} & GPT-4 \citep{openai2023gpt4} (also experiments with GPT-3.5) & PPO \citep{schulman2017proximal} \\ \hline
\citet{song2023self} & GPT-4 \citep{openai2023gpt4} & PPO \citep{schulman2017proximal} \\ \hline
\citet{du2023guiding} & Codex \citep{chen2021evaluating} & DQN \citep{mnih2013playing}, with double Q-learning \citep{vanhasselt2016}, dueling networks \citep{dueling} and multi-step learning \citep{sutton1998} \\ \hline

\citet{quartey2023exploiting} & 
InstructGPT text-davinci-003 (for generating instructions) \citep{ouyang2022training}, Sentence-T5 \citep{ni2021sentencet5} (for encoding object descriptions to vectors) & LPOPL (LTL Progression for Off-Policy Learning) \citep{lpopl} \\ \hline
\citet{reid2022wikipedia} & GPT-2 \citep{Radford2019LanguageMA}, CLIP \citep{radford2021learning}, iGPT \citep{igpt}; & Decision Transformer \citep{chen2021decision}, CQL \citep{kumar2020conservative}, TD3+BC \citep{fujimoto2021minimalist}, BRAC \citep{wu2019behavior}, AWR \citep{peng2019advantageweighted}.\\ \hline
\citet{hu2023language} & GPT-3.5 \citep{gpt35} (text-davinci-003)  & Q-learning \citep{Mnih2015HumanlevelCT} and PPO \citep{schulman2017proximal}\\ \hline

\citet{carta2023grounding} & Flan-T5 780M \citep{flant5} & PPO \citep{schulman2017proximal} \\ \hline
\citet{zhang2023rladapter} & GPT-3.5 \citep{gpt35} and GPT-4 \citep{openai2023gpt4} & PPO \citep{schulman2017proximal} \\ \hline

\end{tabularx}
\end{center}

\end{table*}

%% file: SECTIONS/6_RLPLUSLLM.tex
\section{\texttt{RL+LLM}: Combining Independently Trained Models for Planning} \label{sec:rlplusllm}

The last major class of studies includes those where the RL agent and the LLM are fundamental components of the same framework and where, contrary to the two previous categories, they are independent from each other. In this class, an RL agent is trained to learn specific skills, and the LLM leverages its knowledge of the real world to determine ways to plan over those skills in order to accomplish a task. This combination results in an agent that knows how to perform long-horizon tasks in a specific environment. 


This category can be further refined based on whether planning relies on conversational feedback or not. In the first subcategory, {\texttt{RL+LLM-No Language Feedback}, the LLM generates a static skill graph but does not participate in the planning process after that. In the second subcategory,  ({\texttt{RL+LLM-With Language Feedback}), the user query or the LLM prompt is updated according to the results of the interaction between the agent and the environment in each step. However, we should highlight that {\texttt{RL+LLM-With Language Feedback} studies where the prompt is modified during planning are completely different from the \texttt{RL4LLM-Prompt} studies that were analyzed in section \ref{subsec:RL4LLM-prompt}: The goal of frameworks of the \texttt{RL4LLM - Prompt} category is the improvement of the LLM itself, while the goal of frameworks in the \texttt{RL + LLM} category is planning towards a downstream task that is not directly related to the LLM - or natural language in general.

\subsection{\texttt{RL+LLM-Without Language Feedback}}

Aside from grounding robotic agents to their environment, \texttt{RL + LLM} combinations have been shown to be beneficial for learning multiple, long-horizon tasks in an open-ended environment, as in the case of  \textbf{\citet{yuan2023plan4mc}}, who developed Plan4MC, an framework for executing Minecraft tasks. As the authors explain, exploration under long-horizon tasks owes its difficulty to the size, complexity, and partial observability of open-ended environments. The most common strategy so far in Reinforcement learning literature and practice has been imitation learning, which relies on expert demonstrations or video datasets, which are frequently difficult to obtain. However, even setting aside this obstacle, training an RL agent in a large state space is inevitably hard due to sample inefficiency, while skipping demonstrations is not a viable choice, since it might allow the agent to learn only a very restricted set of skills. As such, the main idea of \cite{yuan2023plan4mc}, it to break down tasks into basic, short-horizon skills, learn those separately, and plan over skills – in other words, find the proper sequence of skills to be executed. In this context, reinforcement learning is used to train the fundamental skills, which are classified as ``Find'', ``Manipulate'', and ``Craft'', independently in advance. It is worth nothing that each type of skill is trained with a different algorithm. The authors prompt ChatGPT \citep{chatgpt} by providing the context, along with an example of the output in the desired format, and ChatGPT produces the skill graph in the desired format, where nodes represent skills and arcs represent ``require'' and ``consume'' relationships. During online planning, the agent alternates between skill planning (i.e., identifying a feasible plan to achieve the goal by performing depth-first search on the skill graph) and skill execution (i.e., selecting policies to solve the complicating skills, and reverting to skill planning if the execution of a task fails). The experimental results confirmed the validity of Plan4MC, which was shown to achieve higher success rate compared to other variations of the algorithm, including those without task decomposition or without separate learning of ``Find''. In a separate set of experiments, ChatGPT was also used to generate the plan, but this model variation was outperformed by the original Plan4MC version.

\subsection{\texttt{RL+LLM-With Language Feedback}}

In a common planning scheme under this category, the LLMs generate instructions for tasks that the agent has already learned through Reinforcement Learning, while the feedback from the environment is used to update the instructions. The feedback from the environment includes natural language (although not necessarily limited to it - see \cite{huang2022inner}) and is used to update the user query or prompt based on the results of the itneraction between the agent and the environment in each step.

``SayCan'' by \textbf{\citet{ahn2022i}} involves a robotic assistant that performs household tasks in a kitchen environment. The authors rely on the observation that, the knowledge of LLMs about the world, while vast, is not physically grounded to the environment that the robotic agent is operating in. As a result, a fully trained robotic agent might not be able to select the appropriate skills to accomplish a task. The role of reinforcement learning in this case is to achieve grounding by helping the agent obtain awareness of the scene. In this case, grounding is measured by calculating the probability of successfully executing a task using a particular skill in a given state. Both the LLM and RL directly participate in computing this probability: the LLM is used to calculate the probability that each skill contributes to completing the instruction, while the affordance function of the RL agent provides the probability that each skill will be executed successfully. The product of those two quantities is the probability that a skill will successfully perform the instruction. Then, the most probable skill is selected, its policy is executed, and the LLM query is amended to include the language description of the skill. The plan is formulated as a dialogue between the robot and the user, where the user provides a high-level instruction and the robot responds by listing the skill sequence that is it going to execute. SayCan was evaluated on 101 different tasks in a real kitchen environment 

To improve the performance of the system, the LLM undergoes prompt engineering to ensure that it produces skill recommendations that are consistent with the user query. In fact, the authors found that the performance of the LLM improved when a) the sequential steps were explicitly numbered, b) the objects mentioned presented variation across the prompt examples, c) careful and error-free phrasing of the names of skills and objects. In a separate set of experiments, SayCan was integrated with Chain of Thought \citep{wei2023chainofthought}, which was shown to improve its performance at negotiation tasks. In addition, it was able to successfully execute instructions provided in languages other than English.

Similarly to``SayCan'', \textbf{\citet{huang2022inner}}proposed Inner Monologue, a framework for planning and interaction with robotic agents that have been trained to execute a variety of skills. Like the previous study of \cite{ahn2022i}, LLMs help the agent understand what the available skills are, how they affect the environment when executed, and how the changes in the environment translate to feedback in natural language. However, contrary to SayCan, Inner Monologue also provides closed-loop feedback to the LLM predictions.

Inner Monologue chains together three components: a) the pre-trained language-conditioned robotic manipulation skills, b) a set of perception models, like scene descriptors and success detectors, and c) human feedback provided by a user that generates natural language instructions towards the robot. The pretrained manipulation skills are short-horizon skills accompanied by short language descriptions, and may be trained through RL. The LLM plays the role of the Planner, whose goal is to find a sequence of skills that achieve the goal expressed by the user. First, the Planner receives the human instruction and breaks it down into a sequence of steps. As it executes the generated plan, the Planner receives three types of textual feedback from the environment: a) Success Detection, which answers whether the low-level skill was successful, b) Passive Scene Description, which is provided without explicitly querying the Planner and includes object recognition feedback and task-progress scene descriptions, and c) Active Scene Description, which is provided by a person or a pretrained model (like a Visual Question Answering model) in response to explicit questions asked by the LLM. As the robot interacts with its environment, the collected feedback is continuously appended to the LLM prompt, thus forming an ``inner monologue'' that closes the loop from the environment to the agent and therefore enhancing the planning capabilities of the LLM. Inner Monologue was tested on simulated and real table-top manipulation environments, as well as a real kitchen mobile manipulation environment. In the latter, pre-trained affordance functions are used for action grounding and the results are compared to SayCan by \cite{ahn2022i} under standard conditions and under adversarial conditions, with added disturbances during control policy executions that cause the plan to fail, In all cases, the embodied feedback provided in the Inner Monologue framework was shown to improve the success rate of the tasks compared to Inner its predecessor, while under adversarial conditions it was only Inner Monologue that was able to consistently complete the instructions successfully. In addition, Inner Monologue was shown to possess significant reasoning capabilities, including continuous adaptation to new instructions, self-proposing goals in cases of infeasiblity, multi-lingual understanding, interactive scene understanding, and robustness to disturbances in human instructions, like swapping the order of feedback or typos. Three failure modes were also observed: False positive and negative success detections, LLM Planning errors due to ignoring the environment feedback, and control errors.

\textbf{\citet{dasgupta2023collaborating}}  proposed a Planner-Actor-Reporter scheme to take advantage of the reasoning capabilities of the LLM and the specialized control skills of a trained RL agent: The LLM acts as the Planner that receives a task description, performs logical reasoning, and decomposes the task into a sequence of instructions that it passes to the Actor, who executes the instructions, while the reporter provides feedback on the action effects back to the Planner.
The framework is implemented in a partially observable 2-D grid-world \citep{pycolab}, with each object possessing a unique combination of color, shape, and texture. Both the Actor and the Reporter are RL agents, trained with VTrace loss. The Actor follows a pre-trained policy that has been trained on simple tasks in the same environment. To allow the agent to receive feedback both from environment observations and through natural language, the policy uses two encoders: a  convolutional visual encoder for visual observations and an LSTM-based language encoder for natural language instructions (e.g., ``Pick up X''), and its action space includes movement within the grid, picking up objects, and examining objects. The Reporter possesses a similar architecture with the Planner, since it also includes encoders for vision and language, but it additionally possesses a memory module and a policy head, which is a binary classifier head that chooses  one of two possible reports. The Reporter observes the results of the Actor's interaction with the environment and communicates with the Planner to inform it of its next command. In every step, the information generated by the Reporter is appended to the dialogue transcript, which is then used as the updated prompt of the Planner, that generates a new instruction for the next step. The authors argue that training the Reporter to ignore noise and produce useful feedback for the planner is more efficient compared to utilizing a large – and therefore expensive – planner. From a robustness perspective, they showed that the framework exhibited robustness for challenging tasks where the Planner needs to explicitly request specific information to incorporate into its next set of instructions, as well as for tasks performed in environments for which the LLM lacks previous semantic experience and therefore needs to perform abstract logical reasoning. They also showed that the Planner-Actor-Reporter scheme is better at learning tasks that are difficult to learn using pure RL Baselines.
Table \ref{tab:rlplusllm-apps} summarizes the tasks and applications of \texttt{RL+LLM} studies.

\input{Supp-Tables/table-rlplusllm-apps}

%% file: Supp-Tables/table-rlplusllm-apps.tex

\begin{table*} 
\caption{RL+LLM Environment and Applications} \label {tab:rlplusllm-apps}
\begin{center}
\renewcommand{\arraystretch}{2}

\scalebox{0.9} {
\begin{tabularx}{1.2\textwidth}{lXX}
\hline

\textbf{RL+LLM Study}   & \textbf{Environment and Long-Term Task} & \textbf{Low-Level Skills} \\ \hline

\citet{yuan2023plan4mc} & Minecraft Tasks (e.g., ``Harvest cooked beef with sword in plains'' & 3 types of skills: Finding, Manipulation, Crafting \\ \hline
\citet{ahn2022i} & 101 real-world robotic in a kitchen. & 551 skills of seven skill families and 17 objects. Skill include include picking, placing and rearranging objects, opening and closing drawers, navigating to various locations, and placing objects in specific configurations. \\ \hline
\citet{huang2022inner} & 3 Families of tasks: Manipulation, Mobile Manipulation, Drawer Manipulation. Mock office kitchen environment with 5 locations and 15 objects. & Navigation and manipulation skills with policies trained from RGB observation.  \\ \hline
\citet{dasgupta2023collaborating} & 2D partially observable grid-world \cite{pycolab} with 4 objects, each of which has a ``secret'' property. Three types of tasks: 1) Secret property conditional task (``If [object 1] is good, pick up [object 1], otherwise pick up [object 2]''. 2) Secret property search task (``The objects are [], [], [], and []. Pick up the object with the good secret property''  & Primitive skills like ``Pick up object X'' or ``Examine object X'' \\ \hline

\end{tabularx}

}
\end{center}

\end{table*}

%% file: SECTIONS/7_DISCUSSION.tex
\section{Discussion} \label{sec:discussion}

\subsection{Goals of Synergy and Reasons for Success}

So far, we have built our taxonomy based on the structural breakdown of the methods analyzed in the studies that combine Reinforcement Learning and Large Language Models, by identifying the way that each of the two models is embedded in the integrated framework where they coexist and potentially interact and pinpointing specific components within this framework where the two model types are interlocked. In this section, we provide a broader view of the studies by examining the goal of this synergy, in line with the inherent features of the two models that make the synergy successful.

\subsubsection{\texttt{RL4LLM}: Responsible AI, Alignment with Human Preference, Performance Improvement}

In the \texttt{RL4LLM} case, particular emphasis is placed on improving the quality of the output of the NLP application that the LLM aims to complete, the performance of the LLM at task execution, or both. Overall, the primary quality considerations in the \texttt{RL4LLM} category are Responsible AI and Alignment with human preferences and intent.

Not surprisingly, given the generative nature of LLMs, Responsible AI is a primary concern of researchers, who wish to ensure the design of models that are not only helpful, but also harmless. Research on mitigating potentially harmful output precedes the era of Large Language Models, with ``harmfulness'' manifesting itself in a variety of ways: offensive responses \cite{perez2022red} (e.g., unkind, with offensive  jokes, or references to morally questionable or sexual desires), data leakage \cite{perez2022red,bai2022constitutional} (i.e., the use of confidential data, such as Social Security Numbers, for training the model, which can then be inferred by an adversary), generated contact information (such as phone numbers, home addresses, and e-mail addresses) \citep{perez2022red}, distributional bias - i.e., text that is negative more often for specific groups \citep{perez2022red,bai2022training}, engagement to sensitive questions \citep{bai2022training,bai2022constitutional}. Notably, all studies which emphasize LLM harmlessness fall under the ``fine-tuning'' umbrella, either with or without human feedback, a fact that indicates that careful prompt design is not always sufficient to guarantee the adherence of the output to responsible AI principles. In fact, the variety of ways in which harmful content can be generated can only be covered through extensive examples, in which few-shot learning is not enough. Naturally, the construction of fine-tuning datasets – which embed the preference and ethical standards of humans – and the subsequent tuning of the model parameters through Reinforcement Learning is a natural choice. Helpfulness, i.e., the alignment between the goals and preferences of the user and the LLM output is another aspect of output quality: For example, an LLM assistant that is tasked to generate Python code is expected to produce clean, executable, and correct results.

\subsubsection{\texttt{LLM4RL}: Efficiency, Grounding, and Human Preferences}

Studies in this class depart from the field of Natural Language Processing and extend to applications where the use of a language model would seem irrelevant in the pre-LLM era. A detailed review of the studies presented in section \ref{sec:rl4llm} reveals that LLMs possess three particular features which make the collaboration between them and RL agents successful:

\begin{enumerate}
\item Ability for zero-shot or few-shot learning: the ability of LLMs to learn through no examples or few examples of desired behavior allows to align their output to human feedback. This alignment is primarily utilized for RL agent reward design (\ref{subsec:reward}) with the goal to generate appropriate reward signals that successfully represent human preferences.
\item Real-world knowledge: LLMs possess vast “knowledge” of the real world, which allows them to explore new behaviors and generate training data. Both capabilities result in time and cost-efficient RL training by a) helping the agent avoid expensive exploration, particularly in open-ended environments, b) eliminating the need for expensive data collection and c) reducing the need for from-scratch training, since they allow policies to be transferred between agents.
\item Reasoning capabilities: for applications involving robotic manipulation, the robotic agent is the one possessing real-world knowledge about its environment, while the LLM is used for grounding the actions of the agent to the environment by ensuring those actions achieve the desired task.
\end{enumerate}

\subsubsection{\texttt{RL+LLM}: Planning}
The goal of all studies in the \texttt{RL+LLM} category is successful planning and execution of relatively complex tasks. In all cases, the agent is equipped with a set of skills that they have already learned through RL. The LLM then helps the agent combine those tasks in order to execute longer-horizon, complex tasks that generally require the execution of more than one of those simple skills, in the correct order. Without the LLM, the agent would have to learn the long-term skills from scratch. However, as we discussed in section \ref{sec:rlplusllm}, training for long-term tasks, especially in complex and partially observable environments, can be data intensive, suffer from sample inefficiency, and eventually unsuccessful. Therefore, instead of explicitly learning complex tasks, planning determines the appropriate sequence of basic skills that have to be executed. LLMs are suitable for planners because they can reason about possible skills execution sequences based on their knowledge of the real world.

\subsection{Shortcomings}

\subsubsection{\texttt{LLM4RL}}

While the available results of the \texttt{LLM4RL} and \texttt{RL+LLM} synergy are impressive and more than promising with regards to the potential for future development, we can identify a set of shortcomings of this class of problems, which refer to two key metrics: the applicability of each framework, and the scalability of the process. 

\paragraph{Applicability.}
Despite the wide applicability of RL agents in domains like (see subsection \ref{subsec:scope} for more details), our review of the work on the \texttt{LLM4RL} and \texttt{RL+LLM} classes reveals that practically all applications of the relevant studies are limited to either benchmarking environments, games, or robotic environments (Tables \ref{tab:llm4rl-apps} and \ref{tab:rlplusllm-apps}), a trend that might initially raise questions about the applicability of the synergy to real-world scenarios beyond household tasks or games. We can attribute this apparent limitation to three reasons: First, the majority of studies presented in sections \ref{sec:rl4llm} and \ref{sec:rlplusllm} focus on introducing novel concepts involving the use of LLMs for tasks that have traditionally been performed otherwise. As proofs-of-concept, they are therefore well-suited to benchmarking environments, like Atari games. Second, before a combined modeling framework is deployed in the real-world, its behavior must be extensively tested for safety, security, and responsible AI considerations.  The amplitude of research on Responsible AI on LLMs, both in the \texttt{RL4LLM} domain and in aside from that, serves as a proof that these considerations are taken seriously by the scientific community. Therefore, it will likely not be long until the \texttt{LLM4RL} classes encompass practical real-world applications of control systems in areas like healthcare and finance. Third, the key strength of LLMs that enables this synergy, i.e., the ability to convey human sense and preferences restricts, at the same time, the range of applications that can be accommodated by \texttt{LLM4RL} frameworks. This limitation applies to both the specification of goals or desired behavior through human language, as well as on the representation of the state using natural language \citep{du2023guiding}.

Even within the realm of the commonly used benchmarking environments, the applicability of \texttt{LLM4RL} methods is often constrained by the limitations of the frameworks themselves. For example, some frameworks, such as the GLAM method by \cite{carta2023grounding} are exclusively limited to textual environments, while the ELLM method by \cite{du2023guiding} assumes a natural language textual representation of the agent’s state. Other methods (e.g., TEXT2REWARD by \cite{xie2023text2reward}  are capable of handling relatively simple tasks, but are yet to be tested for more complex tasks.

\paragraph{Performance.}
Aside from applicability, performance is another parameter that requires further evaluation in \texttt{LLM4RL} studies, with the specific requirements varying among studies. For example, \cite{hu2023language} identify the need to fine-tune InstructRL to improve its performance at test-time.In other cases, the performance of the underlying LLM is shown to be sensitive to prompt choice or even prone to errors despite well-formulated prompts (e.g., \cite{du2023guiding}. Certain language model shortcomings had already been identified prior to the rapid expansion of LLMs - for example, the policy transfer framework of \cite{jiang2023languageinformed} was shown to occasionally suffer from ``catastrophic forgetting'', which significantly reduced the benefits of the agent policy initialization. 

\paragraph{Scalability.}
Finally, the scalability of the solution as the state and action space of the RL agents grows is a potential challenge. As pointed by \cite{carta2023grounding}, scaling up can be computationally inefficient and therefore constrain the application to a single environment and relatively small LLMs.

\subsubsection{\texttt{RL+LLM}}

The combination of RL and LLM for planning long and complex tasks is showing promising results in both studies included in the \texttt{RL+LLM} class. However, an inevitable outcome of such a synergy is that the final model can eventually only be as good as the individual skills for which the agent has been already trained for, irrespective of the robustness of the skill plan generated by the LLM. As pointed in the SayCan paper \citep{ahn2022i}, there is a chance that the system cannot react in case the execution of the intermediate skills fails. Similarly, low success rate of specific individual skills (“Find-Skills”) are the key limitations highlighted by \cite{yuan2023plan4mc}, therefore hindering the end-to-end execution of the plan generated by the Plan4MC method.

\subsection{Alternatives}

Having reviewed the ways that RL and LLM collaborate, along with the strengths and weaknesses of each framework, we are now exploring the existence of LLM-based approaches designed to achieve the same goals without involving RL agents. We investigate the following questions:

\begin{enumerate}
    \item \textit{Is RL required for fine-tuning an LLM?}
    \item \textit{Is RL required for prompt optimization?}
    \item \textit{Is RL required for an LLM to achieve a non-NLP-related task?}
\end{enumerate}

Interestingly, the answer to all the above questions is ``no''. In the discussion that follows, we offer a brief review of state-of-art frameworks that serve as counterexamples. These frameworks are out of the scope of this taxonomy, since they do not rely on the synergy between an RL agent and a LLM - aside, of course, from the use of RLHF for the initial training of the LLM.

\subsubsection{Fine-tuning an LLM without RL: Syndicom, Rain, LIMA}

In section \ref{subsec:finetuning},we presented how RL is used to fine-tune a trained LLM to improve its output for specific tasks. In this section, we present recent studies that achieve fine-tuning without using RL – instead, they use either supervised training methods \citep{zhou2023lima,richardson2023syndicom} or self-evaluation \citep{li2023rain} using specially crafted datasets.

The LIMA (Less Is More for Alignment) model \citep{zhou2023lima} was fine-tuned using supervised learning. The authors analyzed their Superficial Alignment Hypothesis, according to which “a model’s knowledge and capabilities are learnt almost entirely during pretraining, while alignment teaches it which subdistribution of formats should be used when interacting with users”. LIMA creators fine-tune a LLaMa language model with 65 billion parameters using standard supervised loss and a small dataset of 1000 prompt-response pairs. The responses of LIMA outperform  GPT-4\citep{openai2023gpt4}, Bard, and DaVince003 models, based on human evaluation, and demonstrate ability to handle complex queries and generalize well to previously unseen tasks.

In the SYNDICOM framework by \citep{richardson2023syndicom}, the creators fine-tuned a conversational agent to enhance its commonsense reasoning. SYNDICOM consists of two components: First, a dataset containing valid and invalid responses in dialogue contexts, with the invalid ones accompanied by natural language feedback. The authors build a template by randomly sampling from ATOMIC and use GPT-3 to convert the template to natural-sounding dialogue and mark the invalid responses, while human feedback is provided by crowd workers. The second key component of SYNDICOM is a training procedure of a feedback model and a response generation model: First, the feedback model is trained to predict the natural language feedback for invalid responses. Then, the response generation model is trained based on the invalid response, the predicted feedback, and the dialogue. The quality of SYNDICOM responses was shown to outperform ChatGPT based on both ROUGE-1 score and human evaluation.

In a different study, \citep{li2023rain} proposed the RAIN (Rewindable Auto-regressive Inference) method to produce LLM responses aligned to human intent by fine-tuning through self-evaluation and rewind mechanisms. RAIN is a self-alignment model, i.e., does not receive external supervision, and rather allows LLMs to evaluate their output and use the evaluation results to improve it. In a nutshell, RAIN searchers over token sets, with each token set mapping to the node of a search tree. The search consists of an inner and an outer loop. The inner loop, which updates token attributes, consists of a forward search step from root to leaf through heuristic simulation and a backward rewind step to the root. The outer loop adjusts the token set probabilities and determines the next token set. RAIN was shown to outperform LLaMA30B in terms of harmlessness and perform equally well in terms of helpfulness and was also shown to outperform LLaMA-2-chat 13B in terms of truthfulness.

\subsubsection{Prompt Optimization Without RL: Learning-Based Prompt Optimization}
In subsection \ref{subsec:RL4LLM-prompt}, we reviewed studies where RL is used for LLM prompt engineering. Nonetheless, RL is not the sole method for conducting prompt engineering: \citep{sun2023offline} summarized state-of-art methods on learning-based prompt optimization, with examples where prompt optimization is achieved through methods like Beam Search \citep{pryzant2023automatic} or Evolution Strategy \citep{zhou2023large}. However, every single of the \texttt{RL4LLM-Prompt} frameworks presetned in this study was able to overcome traditional challenges that were primarily related to training efficiency of supervised learning methods. RLPrompt \citep{deng2022rlprompt} combined multiple desirable properties which previously had not been present collectively present in any framework: it is automated, gradient-free (therefore eliminating the need to access or compute gradients, which can be computationally expensive), uses frozen LMs (thus not updating any LM parameters), efficient (since it guides optimization through the RL reward information), transferable between different langauge models (due to the use of discrete prompts rather than embeddings), and capable of few-shot and zero-shot learning (since the reward function eliminates the necessity for supervised data). TEMPERA \citep{zhang2022tempera} outperformed RLPrompt in multiple tasks like fine-tuning, prompt tuning and discrete prompt search. Finally, Prompt-OIRL was the first model to address the challenged of inference time evaluation (through offline prompt evaluation) and expensive online prompt optimization (through offline prompt optimization without access to the target LLM).

\subsubsection{LLMs for non-NLP tasks}

As established in section \ref{sec:llm4rl}, integrating a Large Language Model in a RL framework allows us to utilize the vast knowledge and grounding capabilities of LLMs and achieve a variety of control tasks that are not inherently related to natural language, ranging from playing games to robotic manipulation. We also reviewed studies where the output of LLMs together with learned robotic policies can be used for planning or sequential decision-making tasks in the LLM+RL category. Particularly in the realm of robotics, we showed (\ref{sec:llm4rl} and \ref{sec:rlplusllm}) that grounding the agent to the natural environment is a key challenge that LLMs have successfully addressed. 

KOSMOS \citep{huang2023language} is a multimodal Large Language Model that has been trained on web-scale multimodal corpora, including text, image-caption pairs and documents with both images and text. The goal of KOSMOS is to align perception with LLMs, practically allowing models to see and talk. The key idea behind KOSMOS is directly analogous to that of Large Language Models, since it is trained to predict the most likely next token. However, it extends this principle beyond language, showing successful performance on vision tasks as well. More specifically, the model is capable of successfully executing dialogue tasks, visual explanation and Question-Answering, number recognition, and image captioning. 

Similarly, PaLM-E \citep{driess2023palm} is a general-purpose multimodal language model for embodied reasoning, visual-language, and language tasks. Inputs such as images and neural 3D representations are embedded alongside text tokens and passed as input to the Transformer. Incorporating continuous inputs from various sensor modalities of the embodied agent can enable the multimodal language model itself to make grounded inferences for sequential decision making in the real world. PaLM-E transfers knowledge from visual-language domains into embodied reasoning, such as sequential robotic planning and answering questions about the observable world. This knowledge transfer leads to high data efficiency for robotics tasks. PaLM-E operates on multimodal sequences of tokens with inputs such as images and neural 3D representations alongside text tokens. The authors demonstrate that a generalist multi-embodiment agent can be trained leveraging transfer learning across modalities, by incorporating embodied data into the training of a multimodal LLM. Like KOSMOS-1, PaLM-E can perform tasks such as zero shot multimodal chain of thought, visually-conditioned jokes, zero-shot multi-image relationships, spatial grounding, robot visual perception, dialogue and planning etc.

GPT-4V by OpenAI \citep{gpt4vreport} is a multimodal LLM that has been trained to analyze and understand text and image input and generate text outputs demonstrates impressive performance on various tasks, such as exams, logic puzzles, as well as vision and language tasks. GPT-4V was trained on a large-scale corpus of web data, including both positive and negative examples (right and wrong solutions to problems, weak and strong reasoning, self-contradictory and consistent statements) and of various ideologies and ideas. Note that the model’s capabilities seem to come primarily from the pre-training process.

It is interesting to note that multimodality is not necessary for an LLM to succesfully execute non-NLP tasks. A typical example is the SPRING framework \citep{wu2023spring}, where an LLM learns to play complex, open-world games like Crafter or Minecraft by reading the Latex source code of the related academic papers. A directed acyclic graph is constructed, with gameplay specific questions as nodes and dependencies between questions as edges. The experiments demonstrated that the LLM show that when using chain-of-thought prompting, LLMs can successfully execute complex tasks, while SPRING’s zero-shot performance exceeded that of state-of-art RL algorithms for 1 million training steps.

%% file: SECTIONS/8_CONCLUSIONS.tex
\section{Conclusions and Future Work} \label{sec:conclusions}

In this work, we have proposed the RL/LLM Taxonomy Tree, a comprehensive classification of state-of-art computational frameworks that combine Reinforcement Learning Agents and Large Language Models to achieve a target task. We have therefore identified three core classes, namely RL4LLM, which use RL to improve the performance of an LLM; LLM4RL, where an LLM assists the training of an RL agent; and RL+LLM, where an RL agent and an LLM participate in a common framework for planning downstream tasks. We have further divided each class into subcategories based on observations that differentiate the studies that belong in each one. Since each category corresponds to a distinct type of synergy between RL and LLMs, we have explored the key motivations behind the development of the frameworks included in each category and have explained which key strengths of RL and LLMs are utilized each time. The adaptability of RL to NLP tasks thanks to their sequential decision-making nature, as well as the reasoning capabilities and vast knowledge about the real world that LLMs possess serve as testaments for the success of the synergy, resulting in models that are aligned with human intent and Responsible AI principles. In addition, by reviewing the prolific literature on alternative methods, we acknowledge that, for most applications, this synergy is not the only choice. Finally, since LLMs are a relatively new area of Artificial Intelligence, there still exist potential shortcomings; those primarily concern the applicability of LLM4RL and RL+LLM frameworks, along with aspects like computational efficiency and scalability. Nevertheless, the pace of research is so rapid that we can only anticipate substantial improvements in those areas as well.

This review is intended to help researchers understand RL-LLM synergies and develop their own AI frameworks. In the future, we will keep classifying new studies based on the RL/LLM Taxonomy Tree and, if appropriate, expand it to capture novel categories that, given the pace of ongoing research, will almost certainly arise. Undoubtedly, the future holds boundless possibilities for RL-LLM synergies in this regard.